\documentclass[conference]{IEEEtran}

\usepackage{multicol}
\usepackage[numbers]{natbib}
\usepackage{hyperref}
\hypersetup{bookmarks=true}
\usepackage{times}

\usepackage{amsmath} 
\usepackage{amsfonts} 
\usepackage{soul} 
\usepackage{epsfig} 
\usepackage[ruled,vlined]{algorithm2e}
\usepackage[noend]{algpseudocode}
\usepackage[table]{xcolor} 
\usepackage{multirow} 

\usepackage{listings}
\usepackage{color}
\definecolor{codegreen}{rgb}{0,0.5,0}
\definecolor{codeblue}{rgb}{0,0,0.9}
\definecolor{codegray}{rgb}{0.5,0.5,0.5}
\definecolor{codepurple}{rgb}{0.58,0,0.82}
\definecolor{backcolour}{rgb}{0.95,0.95,0.92}
\definecolor{backcolour2}{rgb}{0.9,0.9,0.9}
\definecolor{codered}{rgb}{0.5,0,0}
\definecolor{textcodered}{rgb}{0.4,0,0}
\definecolor{palegray}{rgb}{0.98,0.98,0.99}

\lstdefinestyle{mystyle}{
    backgroundcolor=\color{backcolour},
    commentstyle=\color{codegreen},
    keywordstyle=\color{codeblue},
    numberstyle=\tiny\color{codegray},
    stringstyle=\color{codegreen},
    breakatwhitespace=false,
    breaklines=true,
    captionpos=b,
    keepspaces=true,
    numbersep=5pt,
    showspaces=false,
    showstringspaces=false,
    showtabs=false,
    tabsize=2,
    otherkeywords={with},
    basicstyle=\ttfamily\scriptsize
}

\begin{document}

\title{\textbf{Factory}: Fast Contact for Robotic Assembly}

\author{
\authorblockN{
Yashraj Narang\authorrefmark{1},
Kier Storey\authorrefmark{1},
Iretiayo Akinola\authorrefmark{1}, 
Miles Macklin\authorrefmark{1},
Philipp Reist\authorrefmark{1},
Lukasz Wawrzyniak\authorrefmark{1},
}
\authorblockN{
Yunrong Guo\authorrefmark{1},
Adam Moravanszky\authorrefmark{1},
Gavriel State\authorrefmark{1}, 
Michelle Lu\authorrefmark{1},
Ankur Handa\authorrefmark{1},
Dieter Fox\authorrefmark{1}\authorrefmark{2}
}
\authorblockA{
\authorrefmark{1}NVIDIA Corporation, Santa Clara, USA}
\authorblockA{
\authorrefmark{2}Paul G. Allen School of Computer Science \& Engineering, University of Washington,
Seattle, USA}
}

\maketitle

\begin{abstract}
Robotic assembly is one of the oldest and most challenging applications of robotics. In other areas of robotics, such as perception and grasping, simulation has rapidly accelerated research progress, particularly when combined with modern deep learning. However, accurately, efficiently, and robustly simulating the range of contact-rich interactions in assembly remains a longstanding challenge. In this work, we present \textit{Factory}, a set of physics simulation methods and robot learning tools for such applications. We achieve real-time or faster simulation of a wide range of contact-rich scenes, including simultaneous simulation of $1000$ nut-and-bolt interactions. We provide $60$ carefully-designed part models, $3$ robotic assembly environments, and $7$ robot controllers for training and testing virtual robots. Finally, we train and evaluate proof-of-concept reinforcement learning policies for nut-and-bolt assembly. We aim for \textit{Factory} to open the doors to using simulation for robotic assembly, as well as many other contact-rich applications in robotics. Please see our website for supplementary content, including videos. \footnote{\url{https://sites.google.com/nvidia.com/factory/}}
\end{abstract}

\IEEEpeerreviewmaketitle

\section{Introduction}

Assembly is an essential, but highly challenging area of manufacturing. It includes a diverse range of operations, from peg insertion, electrical connector insertion, and threaded fastener mating (\textit{e.g.}, tightening nuts and bolts), to wire processing, cable routing, and soldering \cite{assembly_magazine_2022, kimble_benchmarking_2020}. These operations are ubiquitous across the automotive, aerospace, electronics, and medical industries \cite{assembly_magazine_2022}. However, assembly has been exceptionally difficult to automate due to physical complexity, part variability, and strict reliability requirements \cite{kimble_benchmarking_2020}.

In industry, robotic assembly methods may achieve high precision, accuracy, and reliability \cite{assembly_magazine_2022, lian_benchmarking_2021, von_drigalski_robots_2019}. However, these methods can be highly restrictive. They often use expensive equipment, require custom fixtures, have high setup times (\textit{e.g.}, tooling design, waypoint definition, parameter tuning) and cycle times, and are sensitive to variation (\textit{e.g.}, part appearance, location). Custom tooling and part-specific engineering are also cost-prohibitive for high-mix, low-volume settings \cite{kimble_benchmarking_2020}. In research, methods for robotic assembly often use less-expensive equipment, require fewer custom fixtures, achieve increased robustness to variation, and may recover from failure \cite{inoue_deep_2017, luo_robust_2021, suarez-ruiz_can_2018}. Nevertheless, these methods often have lower reliability, higher setup times (\textit{e.g.}, demo collection, real-world training, parameter tuning), and/or higher cycle times.

Meanwhile, physics simulation has become a powerful tool for robotics development. Simulators have primarily been used to verify and validate robot designs and algorithms \cite{afzal_study_2020}. Recent research has demonstrated a host of other applications: creating training environments for virtual robots \cite{ehsani_manipulathor_2021, li_igibson_2021, szot_habitat_2021, xiang_sapien_2020}, generating large-scale grasping datasets \cite{eppner_acronym_2021, huang_defgraspsim_2022}, inferring real-world material parameters \cite{matl_inferring_2020, matl_stressd_2020}, simulating tactile sensors \cite{narang_sim-to-real_2021, si_taxim_2021, wang_tacto_2020}, and training reinforcement learning (RL) agents for manipulation and locomotion \cite{chen_system_2021, peng_deeploco_2017}. Compelling works have now shown that RL policies trained in simulation can be transferred to the real world \cite{allshire_transferring_2021, andrychowicz_learning_2020, chebotar_closing_2019, ha_flingbot_2021, mahler_learning_2019, rudin_learning_2021, wu_mat_2019}.

\begin{figure}
    \centering
    \includegraphics[width=\columnwidth]{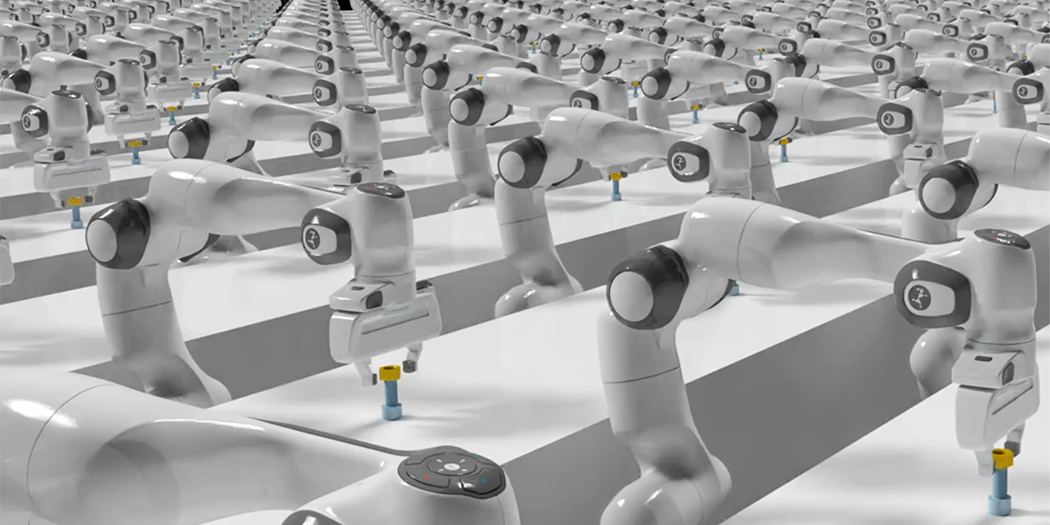}
    \caption{Rendering of Franka robots interacting with nut-and-bolt assemblies in Isaac Gym using methods from \textbf{Factory}. The simulation contains $128$ parallel environments and is executing in real-time on a single GPU.}
    \label{fig:teaser}
\end{figure}

Nevertheless, the power of physics simulation has not substantially impacted robotic assembly. For assembly, a simulator must accurately and efficiently simulate contact-rich interactions, a longstanding challenge in robotics \cite{choi_use_2021, horak_similarities_2019, liu_role_2021, zhang_necessity_2020}, particularly for geometrically-complex, tight-clearance bodies. For instance, consider the canonical nut-and-bolt assembly task. Real-world nuts and bolts have finite clearances between their threads, thus experiencing $6$-DOF relative motion rather than pure helical motion. To simulate real-world motion phases (\textit{e.g.}, initial mating, rundown) and associated pathologies (\textit{e.g.}, cross-threading, jamming) \cite{jia_survey_2019}, collisions between the threads must be simulated. However, high-quality surface meshes for a nut-and-bolt may consist of $10k\mbox{-}50k$ triangles; a naive collision scheme may easily exceed memory and compute limits.
Moreover, for RL training, a numerical solver may need to satisfy non-penetration constraints for $1000$ environments in real-time (\textit{i.e.}, at the same rate as the underlying physical dynamics). Despite the omnipresence of threaded fasteners in the world, no existing simulator achieves this performance.

In this work, we present \textbf{Factory}, a set of physics simulation methods and robot learning tools for such interactions (\textbf{Fig.~\ref{fig:teaser}}). Specifically, we contribute the following:

\begin{itemize}
    \item \textbf{A physics simulation module} for fast, accurate simulations of contact-rich interactions through a novel synthesis of signed distance function (SDF)-based collisions, contact reduction, and a Gauss-Seidel solver. The module is accessible within the PhysX physics engine \cite{nvidia_physx_2022} and Isaac Gym \cite{makoviychuk_isaac_2021}. We demonstrate simulator performance on a wide range of challenging scenes. As an example, we simulate $1000$ simultaneous nut-and-bolt assemblies in real-time on a single GPU, whereas the prior state-of-the-art was a single nut-and-bolt assembly at $\frac{1}{20}$ real-time.
    \item \textbf{A robot learning suite} consisting of a Franka robot and all rigid-body assemblies from the NIST Assembly Task Board $1$ \cite{kimble_benchmarking_2020}, the established benchmark for robotic assembly \cite{von_drigalski_robots_2019}. The suite includes $60$ carefully-designed assets, $3$ robotic assembly environments, and $7$ classical robot controllers. The suite is accessible within Isaac Gym. User-defined assets, environments, and controllers can be added and simulated as desired.
    \item \textbf{Proof-of-concept RL policies} for a simulated Franka robot to solve the most contact-rich task on the NIST board, nut-and-bolt assembly. The policies are trained and tested in Isaac Gym. The contact forces generated during policy execution are compared to literature values from the real world and show strong consistency.
\end{itemize}

We aim for \textbf{Factory} to greatly accelerate research and development in robotic assembly, as well as serve as a powerful tool for contact-rich simulation of any kind.

\section{Related Work}

\subsection{Contact-Rich Simulation}
A longstanding challenge in contact-rich simulation is fast, accurate, and robust contact generation, as well as solution of non-penetration constraints. We have found that achieving such performance requires careful consideration of 1) geometric representations, 2) contact reduction schemes, and 3) numerical solvers. Here we review primary options for each, as well as prior results on a challenging benchmark.

\subsubsection{Geometric Representations}
There are 5 major geometric representations in physics simulation for graphics: convex hulls, convex decompositions, triangular meshes (trimeshes), tetrahedral meshes (tetmeshes), and SDFs (\textbf{Fig.~\ref{fig:geometric_representations}}).

\textbf{Convex hulls} cannot accurately approximate complex object geometries, such as threaded fasteners with concavities. \textbf{Convex decompositions} address this issue by approximating the input shape using multiple convex hulls, generated with algorithms such as V-HACD \cite{mamou_volumetric_2016}. While an improvement on single convex hulls, these decompositions can produce spatial artifacts on complex geometries (\textbf{Fig.~\ref{fig:convex_decomposition}}). Even for perfect decompositions, the number of collision pairs to test during contact generation scales as $\mathcal{O}(n^2)$ (where $n$ is the number of convex shapes), impacting memory and performance. Since contacts are generated between convex shapes, undesirable contact normals can be generated, and snagging may occur.

\textbf{Trimeshes} can provide a near-exact approximation of complex geometries. However, the number of collision pairs for contact generation scales as $\mathcal{O}(n^2)$ (where $n$ is the number of triangles), again impacting memory and performance. In addition, since triangles have zero volume, penetrations can be difficult to resolve, motivating techniques such as boundary-layer expanded meshes \cite{hauser_robust_2016}. \textbf{Tetmeshes} can mitigate such penetration issues, but high quality tetrahedral meshing is challenging. Tetrahedra may have extreme aspect ratios in high-detail areas, leading to inaccurate collision checks.

\textbf{SDFs}, which map points to distance-to-a-surface, can provide accurate implicit representations of complex geometries. They enable fast lookups of distances, as well as efficient computation of gradients to define contact normals. However, using SDFs for collisions requires precomputing SDFs offline from a mesh, which can be time- and memory-intensive. Moreover, collision schemes typically test the vertices of a trimesh against the SDF to generate contacts. For sharp objects, simply sampling vertices can cause penetration to occur, motivating iterative per-triangle contact generation \cite{macklin_local_2020}.

We use discrete, voxel-based SDFs as our geometric representation and demonstrate that they provide efficient, robust collision detection for challenging assets in robotic assembly.

\subsubsection{Contact Reduction}

Contact reduction is a powerful technique from game physics for reducing the total number of contacts without compromising simulation accuracy. A naive contact generation scheme between a nut and bolt may generate $\mbox{$\sim$}1\mathtt{e}6$ contacts; a careful one may generate $\mbox{$\sim$}1\mathtt{e}4$. Excessive contacts can impact both memory and stability, and per-contact memory requirements cannot easily be reduced. Since the rigid-body mechanics principle of equivalent systems dictates that a set of distributed forces can be replaced by a smaller set of forces (constrained to produce the same net force and moment), accurate dynamics can be preserved.

Specifically, contact reduction consists of methods for preprocessing, clustering, and maintaining temporal persistence of contacts. We pay particular attention to \textbf{contact clustering}, which generates bins for contacts, reduces the bins, and reduces the contacts within each bin using heuristics. The two most common heuristics are normal similarity and penetration depth \cite{moravanszky_fast_2004}. \textbf{Normal similarity} assigns contacts with similar surface normals to the same bin, culls similar bins, and culls similar contacts, and is often implemented with $k$-means or cube map algorithms \cite{Otaduy2006, rocchi2016, GSMDO12}. \textbf{Penetration depth} culls bins and contacts with negligible penetration, as these often have minimal impact on dynamics. \textbf{Data-driven} methods train networks to perform the reduction \cite{kim2019}, but require a separate data collection and learning phase to be effective.

We combine normal similarity, penetration depth, and an area-based metric to reduce contacts and demonstrate the resulting dynamics across numerous evaluation scenes.

\subsubsection{Solvers}

There are 4 major options for solvers in physics simulation: direct, conjugate gradient/conjugate residual (CG/CR), Jacobi, and Gauss-Seidel. We briefly review these solution methods below; see \cite{andrews2021} for a detailed treatment.

\textbf{Direct solvers}, which may rely on matrix pivoting, do not scale well to large sets of contact constraints, such as the $\mbox{$\sim$}1\mathtt{e}4$ constraints between a nut and bolt mesh. \textbf{CG/CR methods} can handle larger sets of constraints, but are still unable to achieve real-time performance for complex scenarios. \textbf{Gauss-Seidel solvers} are robust and converge quickly for well-conditioned problems, but perform serial operations that do not scale well to large sets of constraints. \textbf{Jacobi solvers} perform parallel operations that can leverage GPU acceleration, but converge slower than the aforementioned techniques.

A naive implementation of Jacobi can outperform Gauss-Seidel for simulating numerous contact-rich interactions. However, we show that contact reduction can greatly accelerate Gauss-Seidel, achieving better performance than Jacobi.

\subsubsection{Benchmark Problem}

\begin{figure}
    \centering
    \includegraphics[width=\columnwidth]{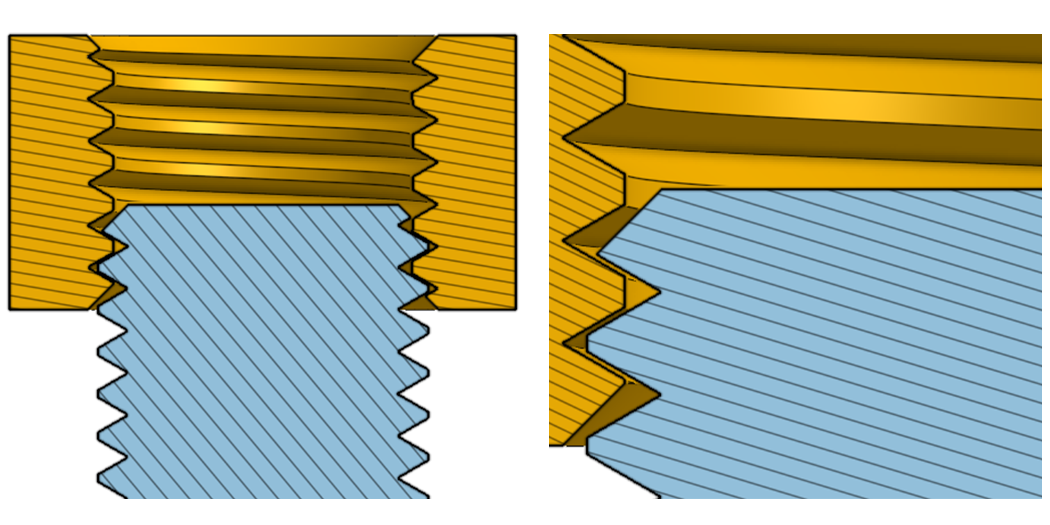}
    \caption{Cross-sectional view of an aligned M16 nut-and-bolt assembly. Nuts and bolts have finite clearances between their threads, thus experiencing 6-DOF kinematics with multi-surface frictional contact, rather than pure helical motion. As follows, a nut can roll and pitch on a bolt, and may experience pathologies such as cross-threading and jamming \cite{jia_survey_2019}. Left: a \textit{loose} model, where the clearances between threads are at their maximum allowable value (ISO 965). Right: zoom  view. Thread roots are not modeled.}
    \label{fig:nut_bolt_threads}
\end{figure}

We pay special attention to the problem of accurately and efficiently simulating a nut-and-bolt assembly (\textbf{Fig.~\ref{fig:nut_bolt_threads}}). In the simulation community, this problem has emerged as a canonical challenge in contact-rich simulation. Moreover, in robotic assembly, tightening nuts onto bolts is critically important, as $\approx 40\%$ of all mechanical assembly operations involve screws and bolts \cite{martin-vega_industrial_1995}.

The 3D finite element method (FEM) is the gold standard for accurate simulation of nut-and-bolt models, capturing deformation phenomena such as pretension (\cite{kim_finite_2007}). However, FEM simulation of a single nut-bolt pair may take $\mbox{$\sim$}1\mbox{-}10~min$ to execute on CPU. As our aim is real-time (or faster) simulation to enable learning methods such as RL, we focus on rigid-body simulation prevalent in graphics and robotics.

In \cite{xu_implicit_2014}, rigid-body contact was simulated using semi-implicit integration, symbolic Gaussian elimination, analytical contact gradients, and an SVD solver. A dynamic scene was simulated of screwing a hex bolt into a threaded hole at $\frac{1}{460}$ real-time ($6.0~h$ for $47~s$). The proposed method was compared against Bullet (2012) \cite{coumans_pybullet_2022}, and a $7\mathtt{e}4\times$ speed-up was measured for a stable quasi-static version of the scene. In \cite{gissler_interlinked_2019}, rigid-body contact was simulated using a smoothed particle hydrodynamics solver that samples rigid surfaces with particles, models contacts as density deviations, and computes pressure-based contact forces implicitly. A quasi-static nut-and-bolt scene was simulated at $\frac{1}{20}$ real-time ($81~s$ for $4~s$). The proposed method was also compared against Bullet (2018), and a $9.5\times$ speed-up was measured for a stable scene. 

In \cite{ferguson_intersection-free_2021}, rigid-body contact was simulated using an extension of incremental potential contact (IPC) \cite{li_incremental_2020} and continuous collision detection (CCD) for curved trajectories. A dynamic, frictionless nut-and-bolt scene was simulated at $\frac{1}{350}$ real-time ($3.5~s$ for $\Delta t=0.01~s$). The proposed method was compared against Bullet (2019), MuJoCo \cite{todorov_mujoco_2012}, Chrono \cite{tasora_chrono_2016}, and Houdini RBD \cite{sidefx_houdini_2022}. Instabilities were observed with Bullet and MuJoCo, and interpenetrations were observed with Chrono and Houdini. In \cite{son_sim-to-real_2020}, rigid-body contact was simulated using a configuration-space contact model. A single large M48 ($48~mm$ diameter) nut-and-bolt was simulated at $4\times$ real-time; however, data and evaluations were limited. Informally, an industry group simulated a dynamic nut-and-bolt scene at $\mbox{$\approx$}\frac{1}{20}$ real-time with variational integrators \cite{bodin_twitter_2020, lacoursiere_ghosts_2007}.

From the previous works, we define the established state-of-the-art for a general-purpose physics simulator that can simulate a nut-and-bolt assembly to be $\frac{1}{20}$ real-time for a single nut-bolt pair. Although fast, such speeds are not optimal for applications such as RL, which benefit from simulation of $\mbox{$\sim$}1\mathtt{e}2\mbox{--}1\mathtt{e}3$ contact-rich interactions in real-time. In this work, we demonstrate real-time simulation of $1000$ nuts-and-bolts, as well as exceptional speeds on other challenging scenes.

\subsection{Robotic Assembly Simulation}
Here we review previous efforts to simulate robotic assembly, as well as efforts to use these simulators and/or real-world training for RL. For another recent review, see \cite{stan_reinforcement_2020}.

\subsubsection{Simulation}
Research efforts such as \cite{xu_implicit_2014, gissler_interlinked_2019, ferguson_intersection-free_2021} have demonstrated advances in contact-rich simulation, often in comparison to physics engines used in robotics. However, the majority of robotics studies use PyBullet, MuJoCo, or Isaac Gym, as they provide robot importers, simulated sensors, parallel simulation, and learning-friendly APIs. Consequently, existing work in simulation for robotic assembly is largely limited to the performance of these simulators.

Efforts using MuJoCo or the robosuite extension \cite{zhu_robosuite_2020} include \cite{davchev_residual_2021, fan_learning_2019, hebecker_towards_2021, jin_trajectory_2021, khader_stability-guaranteed_2021, lee_ikea_2021, schoettler_meta-reinforcement_2020, spector_deep_2020, vecerik_leveraging_2018, vecerik_practical_2019,  vuong_learning_2021, zachares_interpreting_2021, zhang_learning_2021, zhao_towards_2020}. Simulated rigid-body assembly tasks are limited to peg-in-hole insertion of large pegs with round, triangular, square, and prismatic cross-sections, lap-joint mating, and one non-convex insertion \cite{hebecker_towards_2021}. Furthermore, clearances between rigid parts are typically $0.5\mbox{-}10~mm$, substantially greater than real-world clearances.
Efforts using PyBullet include  \cite{apolinarska_robotic_2021, luo_dynamic_2019, shao_learning_2020, yu_roboassembly_2021}. Simulated tasks are again limited to large peg-in-hole insertion and lap-joint mating, with clearances of $\mbox{$\sim$}1~mm$. A handful of research efforts have used other off-the-shelf simulators \cite{beltran-hernandez_variable_2020, hou_data-efficient_2021, lee_making_2020, wu_learning_2021, zhao_towards_2020} and custom simulators \cite{clever_assistive_2021, luo_learning_2021, son_sim-to-real_2020}. Only \cite{son_sim-to-real_2020} simulated a nut-and-bolt, which was exceptionally large ($48~mm$ diameter) with inherently-larger clearances.

From the previous works, we conclude that there have been few successful efforts to simulate assembly tasks with realistic scales, realistic clearances (\textit{e.g.}, $\mbox{$\sim$}0.1~mm$ diametral clearance for a $4~mm$ peg), and complex geometries (\textit{e.g.}, nuts-and-bolts, electrical connectors) within a robotics simulator. In this work, we build a module for PhysX and Isaac Gym that can successfully simulate all rigid components on NIST Task Board $1$ \cite{kimble_benchmarking_2020} with accurate models and real-world clearances.

\subsubsection{Reinforcement Learning}
The studies in the previous subsection, as well as a small number of studies that trained RL for robotic assembly purely in the real world, can be categorized according to their choice of RL algorithm.

Several earlier works used model-based RL algorithms or proposed variants, which explicitly predict environment response during policy learning and/or execution. These efforts leveraged guided policy search \cite{luo_deep_2018, thomas_learning_2018} and iterative LQG \cite{luo_reinforcement_2019}. Such algorithms are sample-efficient, but difficult to apply to contact-rich tasks due to highly discontinuous and nonlinear dynamics and unknown material parameters \cite{spector_deep_2020}.

Most recent works have used model-free, off-policy RL algorithms or variants, which do not predict environment response, and update an action-value function independently of the current policy (\textit{e.g.}, using a replay buffer). These studies have applied Q-learning \cite{inoue_deep_2017}, deep-Q networks \cite{zhang_learning_2021}, deep deterministic policy gradients (DDPG) \cite{apolinarska_robotic_2021, luo_learning_2021, luo_robust_2021, vecerik_practical_2019}, soft-actor critic \cite{beltran-hernandez_variable_2020}, probabilistic embeddings \cite{schoettler_meta-reinforcement_2020}, and hierarchical RL \cite{hou_data-efficient_2021}. These algorithms are typically chosen for sample efficiency, but are often brittle and slow/unable to converge. 

Several other studies use or develop off-policy RL algorithms that leverage human demonstrations, as well as motion planners and trajectory optimizers. These efforts have used residual learning from demonstration \cite{davchev_residual_2021}, guided DDPG \cite{fan_learning_2019}, DDPG from demonstration (DDPGfD) \cite{luo_robust_2021, luo_dynamic_2019, vecerik_leveraging_2018}, and inverse RL \cite{wu_learning_2021}. Notably, \cite{luo_robust_2021} used DDPGfD to achieve state-of-the-art performance in the real world for $3$ insertion tasks from NIST Task Board $1$. They used human demonstrations, real-world training, and human on-policy corrections, achieving a $99.8\%$ success rate over $13k$ trials. Of course, these methods also require demonstration collection or effective planners/optimizers, and are typically limited to the performance of these structured priors.

Finally, a handful of research efforts have successfully used on-policy algorithms or variants. These studies have applied proximal policy optimization (PPO) \cite{hebecker_towards_2021, son_sim-to-real_2020, vuong_learning_2021}, trust region policy optimization \cite{lee_making_2020}, asynchronous advantage actor-critic \cite{shao_learning_2020}, and additional algorithms \cite{khader_stability-guaranteed_2021}. These algorithms are typically stable, easy-to-use, and achieve high return, but are highly sample-inefficient and require long wall-clock time.

We take inspiration from the performance of \cite{luo_robust_2021}, but aim to achieve such performance in a fundamentally different way. We build a module for contact-rich simulation that can help enable roboticists to perform more complicated tasks (\textit{e.g.}, nut-and-bolt assembly) with tight clearances; leverage performant and stable on-policy RL algorithms with high parallelization; avoid tedious human demonstrations and corrections; and mitigate the need for time-consuming (\textit{e.g.}, $\mbox{$\sim$}50$ hours of data in \cite{luo_robust_2021}), costly, and dangerous real-world RL training.

\section{Contact-Rich Simulation Methods}

In this work, we first build a module for PhysX \cite{nvidia_physx_2022} for efficient and robust contact-rich simulation. Specifically, we uniquely combine SDF collisions \cite{macklin_local_2020}, contact reduction \cite{moravanszky_fast_2004}, and a Gauss-Seidel solver \cite{macklin_small_2019}, allowing us to simulate interactions of highly-detailed models substantially faster than previous efforts. We describe methods and results below.

\subsection{SDF Collisions}

An SDF is a function $\phi(\boldsymbol{x}): \mathbb{R}^3 \rightarrow \mathbb{R}$ that maps a point $\boldsymbol{x}$ in Cartesian space to its Euclidean distance to a surface. The surface is implicitly defined by $\phi(\boldsymbol{x}) = 0$, and the sign indicates whether $\boldsymbol{x}$ is inside or outside the surface. The gradient $\nabla \phi(\boldsymbol{x})$ provides the normal at a point $\boldsymbol{x}$ on the surface. Collectively, the SDF value and gradient define a vector to push a colliding object out of the surface.

We generate the SDF for an object via sampling and compute gradients via finite-differencing. Specifically, given a triangle mesh representing the boundaries of the object, we generate an SDF for the mesh at initialization time and store it as a 3D texture, which enables GPU-accelerated texture fetching with trilinear filtering and extremely fast $\mathcal{O}(1)$ lookups. Since our shapes contain many small features, we typically use SDF resolutions of $256^3$ or greater.

To generate an initial set of contacts, we use the method of \cite{macklin_local_2020}, which generates one contact per triangle-mesh face. The contact position on each face is determined by performing iterative local minimization to find the closest point on the face to the opposing shape, using projected gradient descent with adaptive stepping. As an example, detailed M4 nut-and-bolt meshes generate $16k$ contacts; with the above method, these contacts can be generated in $<1~ms$, orders of magnitude faster than typical approaches for convex or mesh collision. 

\subsection{Contact Reduction and Solver Selection}

To motivate our contact reduction methods, we begin with a brief discussion and first-order analysis of the tight coupling between contact generation and solver execution.

In a typical contact generation scheme, each contact only requires approximately $7$ floats (point, normal vector, and distance) and $2$ integers (rigid body indices), for a total of $36$ bytes. However, the memory required to store constraints associated with these contacts is substantially greater; in our implementation, storing a contact and its constraints requires approximately $160$ bytes. Thus, simulating an M4 nut-bolt pair with $16k$ contacts requires approx. $2.5~MB$ per timestep.

For a Jacobi solver with $\Delta t = \frac{1}{60} s$, we require $8$ substeps and $64$ iterations for stable simulation. Thus, memory bandwidth requirements for $16k$ contacts are approximately $1.28~GB$ per frame and $76.8~GB$ per second. Using a state-of-the-art GPU, we can only simulate $20$ nut-and-bolt assemblies in parallel (\textbf{Table~\ref{tab:solver_comparison}}). Unfortunately, although a Gauss-Seidel solver converges faster (\textit{i.e.}, requires fewer substeps and iterations), it would be unreasonably slow for so many contacts due to its inherently serial nature. As reducing per-contact memory can be challenging, contact reduction becomes a compelling strategy for reducing memory and increasing parallelization.

If we can reduce the number of contacts to $2\%$ (\textit{i.e.}, from $16k$ to $300$), the simulation now requires approximately $48~kB$ per timestep. The Jacobi solver now only requires $24~MB$ per frame and $1.44~GB$ per second. We can now hypothetically simulate a maximum of $1100$ nut-and-bolt assemblies in real-time. Solving contact constraints is no longer a performance bottleneck, and we can achieve a level of parallelization suitable for training on-policy RL algorithms. 

In addition, given the far fewer contacts, it is now feasible to use a Gauss-Seidel solver. With $\Delta t = \frac{1}{60} s$, we require only $1$ substep and $16$ iterations for stable simulation. Thus, we can now comfortably simulate $>1000$ nut-and-bolt assemblies in real-time. We demonstrate exactly such performance later. 

\subsection{Implementation of Contact Reduction}

To implement contact reduction, we use the concept of \textit{contact patches}, which are sets of contacts that are proximal and share a normal. We generate contact patches in $3$ phases (\textbf{Algorithm~\ref{alg:collision}}). First, we \textbf{Generate} candidate contacts using SDF collisions for a batch of triangles (size $1024$). Second, we \textbf{Assign} candidates to existing patches in the shape pair based on normal similarity. Third, for unassigned candidates, we \textbf{FindDeepest} (\textit{i.e.}, find the one with deepest penetration), create a new patch, \textbf{BinReduce} (\textit{i.e.}, assign remaining candidates to this patch based on normal similarity), and \textbf{AddPatch} to our list of patches. We repeat until no candidates remain. When performing \textbf{AddPatch}, we also check if a \textit{patch} with a similar normal exists. We either add both patches or replace the existing patch, using a measure that maximizes patch surface area, prioritizes contacts with high penetration depth, and restricts the number of patches to $N$ (where $128 \leq N \leq 256$).

The preceding contact reduction process is performed exclusively in GPU shared memory. Notably, contact reduction does not make contact generation slower; to the contrary, it makes generation faster, as the contact generation kernel does not need to write extensive amounts of data to global memory.

Applying the above procedure to the M4 nut-and-bolt interactions, we reduce the number of contacts from $16k$ to $300$ (\textbf{Fig.~\ref{fig:contact_reduction}}), allowing us to simulate $1024$ assemblies in real-time on an NVIDIA A5000 GPU. Generating and reducing contacts takes $11~ms$, and solving contact constraints takes an additional $3~ms$. A number of additional evaluations follow.

\begin{algorithm}
\SetAlgoLined
\ForEach{shape pair $(a, b)$}{
    $patches = \emptyset$ \\
    \ForEach{batch in triangles($a$, $b$)}{
        $candidates \leftarrow \text{GenerateContacts}(batch)$ \\
        $patches \leftarrow \text{Assign}(candidates, patches)$ \\
        \While{candidates $\neq \emptyset$}{
            $best \leftarrow \text{FindDeepest}(candidates)$ \\
            $patch \leftarrow \text{BinReduce}(candidates, best)$ \\
            $patches \leftarrow \text{AddPatch}(patch, patches)$ 
        }
    }
}
\caption{\textbf{Collision Detection}: An overview of our collision detection pipeline. The input is two potentially contacting shapes ($a$, $b$). The output is a list of patches (\textit{i.e.}, sets of contacts) for each shape pair.}
\label{alg:collision}
\end{algorithm}

\subsection{Performance Evaluations}

\begin{figure}
    \centering
    \includegraphics[width=\columnwidth]{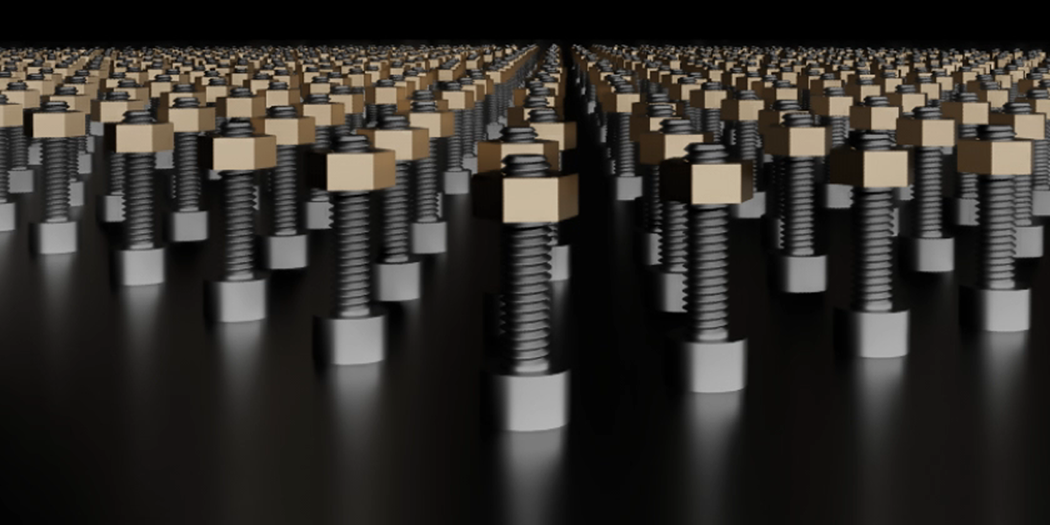}
    \caption{Rendering of the \textbf{M16 nut-and-bolt assemblies} scene, consisting of $1024$ parallel nut-and-bolt interactions executing in real-time.}
    \label{fig:nuts-and-bolts}
\end{figure}

\begin{figure}
    \centering
    \includegraphics[width=\columnwidth]{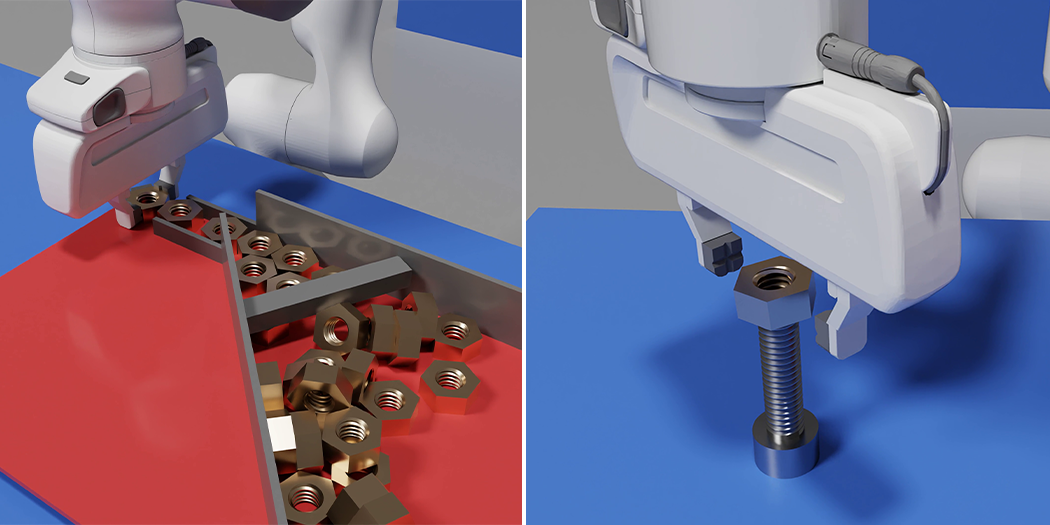}
    \caption{Rendering of the \textbf{Franka robot + M16 nut-and-bolt assemblies} scene, consisting of $128$ parallel Franka robots retrieving nuts from a vibratory feeder mechanism (left) and tightening them onto a bolt (right).}
    \label{fig:vibratory_scene}
\end{figure}

We test our collision detection, contact generation, contact reduction, and solution pipeline on $8$ contact-rich scenes. These scenes were designed to represent a broad range of challenging real-world scenarios, including complex geometries, robot-object interactions, tight clearances, $100$s of interacting bodies, and multi-part mechanisms. The scenes are as follows:
\begin{itemize}
    \item 1024 parallel $\boldsymbol{4\mbox{-}mm}$ \textbf{peg-in-hole assemblies} from the NIST board with ISO-standard clearances ($0.104~mm$).
    \item $1024$ parallel \textbf{M16 nut-and-bolt assemblies} with ISO-standard clearances (\textbf{Fig~\ref{fig:nuts-and-bolts}}). For ease of contact profiling, the coefficient of friction is reduced to $0.05$, allowing the nuts to rotate on the bolts under gravity.
    \item $1024$ parallel VGA-style \textbf{D-subminiature (D-sub) connectors} from the NIST board (\textbf{Fig~\ref{fig:nist_dsub_gears}}). For ease of contact profiling, a $2\mathtt{e}\mbox{-}6~m$ clearance is introduced, allowing the plug-and-socket to mate under gravity.
    \item $1024$ parallel $2$-stage \textbf{gear assemblies} from the NIST board (\textbf{Fig~\ref{fig:nist_dsub_gears}}). An external torque is applied to the intermediate gear to rotate the adjacent gears.
    \item $1024$ \textbf{M16 nuts}, falling into a pile in one environment (\textbf{Fig.~\ref{fig:demos}}).
    \item $1024$ \textbf{bowls} (akin to \cite{xu_implicit_2014}), falling into a pile
    in one environment. To enable larger timesteps while maintaining accuracy, a tiny $1\mathtt{e}\mbox{-}6~m$ \textit{negative} clearance is added.
    \item $1024$ \textbf{toruses}, falling into a pile in one environment.
    \item $128$ parallel \textbf{Franka robot + M16 nut-and-bolt assemblies}. A vibratory feeder mechanism feeds an aggregate of nuts into a channel. The robot grasps the nut from the channel and tightens it onto the bolt using an inverse kinematics (IK) controller (\textbf{Fig.~\ref{fig:vibratory_scene}}, \textbf{App.~\ref{sec:app_additional_scenes}}).
\end{itemize}

\textbf{Table~\ref{tab:simulation_geometric_representations}} describes the geometric representations used in each of the scenes, including SDF resolution and number of triangles. \textbf{Table~\ref{tab:simulation_contact_generation}} provides statistics on contacts before and after contact reduction, as well as timing for reduction and solution. Finally, \textbf{Table~\ref{tab:simulation_performance}} provides simulation performance statistics, including comparisons to real-time. We also qualitatively demonstrate an additional test scene: a \textbf{Franka robot + M16 nuts + flange assembly} scene (\textbf{App.~\ref{sec:app_additional_scenes}}). 

\begin{table*}
\centering
\begin{tabular}{llllllll}
 & \multicolumn{3}{c}{Contact Stats (Before)} & \multicolumn{1}{c}{Contact Handling} & \multicolumn{2}{c}{Contact Stats (After)}& \multicolumn{1}{c}{Contact Solution} \\ \hline
\textbf{Scene} & \textbf{Contacts} & \textbf{Per Pair (avg)} & \textbf{Per Pair (max)} & \textbf{Time}  & \textbf{Per Pair (avg)} & \textbf{Patches} & \textbf{Time} \\ \hline
\rowcolor{gray!10} Peg-in-hole & 5.89e5 & 576 & 576 & 2 ms & 46 & 11 & 1 ms \\ 
Nut-and-bolt & 1.73e7 & 16930 & 16930 & 11 ms & 195 & 53 & 3 ms \\ 
\rowcolor{gray!10} D-sub connector & 1.20e7 & 11746 & 11746 & 12.5 ms & 175 & 36 & 1.5 ms \\ 
Gear assembly & 7.26e7 & 14172 & 31568 & 39 ms & 83 & 26 & 3 ms \\ 
\rowcolor{gray!10} Nuts & 1.81e6 & 2516 & 8304 & 10 ms & 99 &  46 & 2.1 ms \\ 
Bowls & 5.23e5 & 731 & 1160 & 2 ms & 66 & 18 & 4.3 ms \\ 
\rowcolor{gray!10} Toruses & 4.00e5 & 185 & 864 & 4.6 ms & 44 &  20 & 2.8 ms \\ 
Franka + nut-and-bolt & 4.64e5 & 9285 & 10031 & 1.7 ms & 147 &  40 & 2.7 ms \\ 
\end{tabular}
\caption{\label{tab:simulation_contact_generation}Contact statistics. \textit{Contacts} is the total number of contacts in the scene. \textit{Contact Handling} includes pair finding, contact generation, and contact reduction. \textit{Contact Solution} includes contact constraint preparation, iterative constraint solution, and numerical integration.}
\end{table*}

\begin{table*}
\centering
\begin{tabular}{lllllll}
 & \multicolumn{3}{c}{Timestepping} & \multicolumn{2}{c}{Simulation Stats} \\ \hline
\textbf{Scene} & \textbf{Substeps} & \textbf{Pos Iterations} & \textbf{Vel Iterations} & \textbf{Time} & \textbf{Real-time} \\ \hline
\rowcolor{gray!10} Peg-in-hole & 1 & 4 & 1 & 3 ms & 5689x \\ 
Nut-and-bolt & 1 & 20 & 1 & 14 ms & 1219x \\ 
\rowcolor{gray!10} D-sub connector & 4 & 4 & 1 & 14 ms & 305x \\ 
Gear assembly & 4 & 4 & 1 & 42 ms & 102x \\ 
\rowcolor{gray!10} Nuts & 1 & 16 & 1 & 12 ms & 1.39x \\ 
Bowls & 2 & 50 & 1 & 6.3 ms & 1.32x \\ 
\rowcolor{gray!10} Toruses & 1 & 16 & 1 & 7.4 ms & 2.25x \\ 
Franka + nut-and-bolt & 4 & 16 & 1 & 4.4 ms & 121x \\ 
\end{tabular}
\caption{\label{tab:simulation_performance}Performance statistics. The baseline timestep size (before substepping) is $\frac{1}{60}~s$. \textit{Pos Iterations} and \textit{Vel Iterations} denote the number of position and velocity iterations in the Gauss Seidel solver. \textit{Time} denotes total simulation time per frame, which is the sum of \textit{Contact Handling} and \textit{Contact Solution} from \textbf{Table~\ref{tab:simulation_contact_generation}}. \textit{Real-time} denotes stable timestep size (after substepping) relative to simulation time, scaled by the number of parallel environments.}
\end{table*}

Although we defer to the tables for complete performance assessments, key observations include the following:

\begin{itemize}
    \item Contact reduction can reduce contact counts by over $2$ orders-of-magnitude compared to naive methods.
    \item Contact handling time (\textit{i.e.}, pair finding, generation, reduction) is typically dominant compared to solution time.
    \item Parallelization achieves a $3$-orders-of-magnitude speed-up over real-time single-threaded computation.
\end{itemize}

\section{Robot Learning Tools}

We have thus far evaluated our physics simulation module over a diverse array of contact-rich scenes. However, the module is an extension of PhysX \cite{nvidia_physx_2022}. For convenient use in robot learning, we have integrated our module into Isaac Gym \cite{makoviychuk_isaac_2021}, which can use PhysX as its physics engine. To use our contact methods for arbitrary assets, the user simply has to include an $\mbox{$<$}sdf\mbox{$>$}$ tag in URDF descriptions (\textbf{Listing~\ref{lst:nut_urdf}}).

For applications to robotic assembly, assets and scenes related to NIST Task Board $1$ may be particularly useful. Thus, we provide 1) $60$ assets from the NIST board, 2) $3$ robotic assembly scenes for RL training in Isaac Gym, and 3) $7$ classical robot controllers in Isaac Gym to accelerate learning. Here we describe our assets, environments, and controllers. 

\subsection{Assets}

The NIST Task Board $1$ consists of $38$ unique parts. However, the CAD models publicly provided for these parts are not suitable for high-accuracy physics simulation. In particular, the models for the nuts, bolts, pegs, and gear assembly do not conform to real-world tolerances and clearances; in assembly, mating parts together with tight clearances is precisely the most significant challenge. Furthermore, the models for the electrical connectors were sourced from public repositories rather than manufacturers (\textit{e.g.}, the D-sub plug and socket), were geometrically incompatible (\textit{e.g.}, the RJ45 plug and socket, which interpenetrate), were incomplete (\textit{e.g.}, the USB socket, which lacks mating features), and/or were designed using hand measurements (\textit{e.g.}, the Waterproof plug and socket). Regardless of simulator accuracy, inaccurate geometries (esp. interpenetration) will lead to unstable or inaccurate dynamics.

We provide $60$ high-quality, simulation-ready part models, each with an Onshape CAD model, one or more OBJ meshes, a URDF description, and estimated material properties (\textbf{Table~\ref{tab:assets_standard}} and \textbf{Table~\ref{tab:assets_connectors}}). These models include all the parts on the NIST Task Board $1$, as well as dimensional variations. The assets for the nuts, bolts, pegs, and gearshafts conform to ISO $724$, ISO $965$, and ISO $286$ standards and contain \textit{loose} and \textit{tight} configurations that correspond to the extremes of the tolerance band. The CAD models for the electrical connectors were sourced from manufacturer-provided models. Each part of each connector contains a \textit{visual mesh}, directly exported from the CAD models, and a \textit{collision mesh}, carefully redesigned to simplify external geometry while faithfully preserving mating features (\textit{e.g.}, pins and holes).

\begin{figure}
    \centering
    \includegraphics[width=\columnwidth]{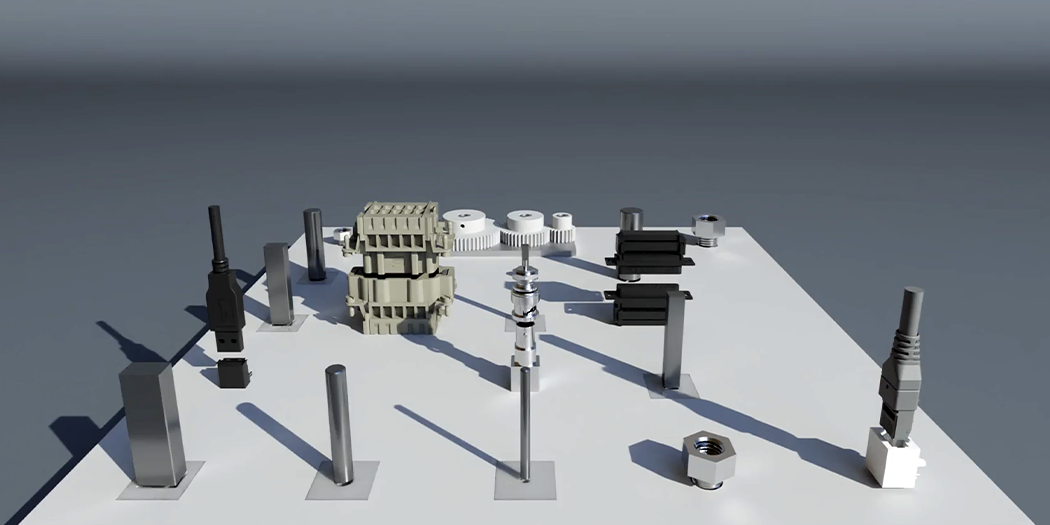}
    \caption{Rendering of a simulated NIST Task Board $1$, demonstrating the provided assets. We provide simulation and RL training environments for all rigid components of the board. Compare to the real board in \textbf{Fig.~\ref{fig:nist_board_real}}.}
    \label{fig:nist_board_sim}
\end{figure}

\subsection{Scenes}\label{sec:scenes}

\begin{figure}
    \centering
    \includegraphics[width=\columnwidth]{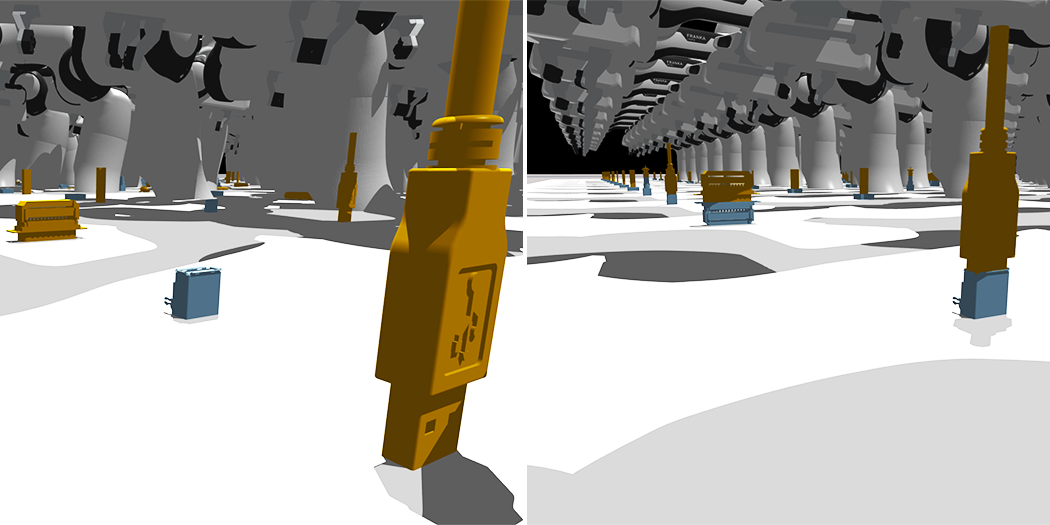}
    \caption{Visualization of \textbf{FrankaInsertionEnv}. Each environment consists of a Franka robot and an insertion assembly from NIST Task Board $1$. Left: The default initial state, where the positions of the parts are randomized on the work surface. Right: The default goal state, where all parts are inserted.}
    \label{fig:franka_insertion_env}
\end{figure}

\begin{figure}
    \centering
    \includegraphics[width=\columnwidth]{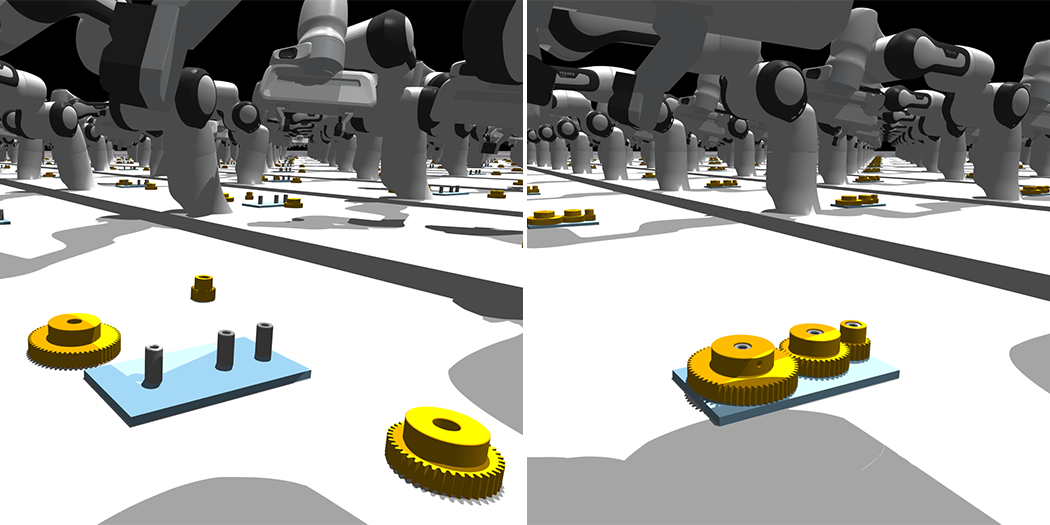}
    \caption{Visualization of \textbf{FrankaGearsEnv}. Each environment consists of a Franka robot and the gear assembly from NIST Task Board 1. Left: The default initial state, where the positions of the gears are randomized on the work surface. Right: The default goal state, where all gears are aligned.}
    \label{fig:franka_gears_env}
\end{figure}

We provide $3$ robotic assembly scenes for Isaac Gym that can be used for developing planning and control algorithms, collecting simulated sensor data for supervised learning, and training RL agents. Each scene contains a Franka robot and disassembled assemblies from NIST Task Board 1. All scenes have been tested with up to $128$ simultaneous environments on an NVIDIA RTX 3090 GPU. The scenes are as follows:
\begin{itemize}
    \item \textbf{FrankaNutBoltEnv}, which contains a Franka robot and nut-and-bolt assemblies of the user's choice (M4, M8, M12, M16, and/or M20). The nuts and bolts can be randomized in type and location across all environments. The default goal is to pick up a nut from a work surface and tighten it to the bottom of its corresponding bolt. Our own RL training results on this environment will be discussed in detail in the next section.
    \item \textbf{FrankaInsertionEnv}, which contains a Franka robot and insertion assemblies of the user's choice (round and/or rectangular pegs-and-holes; BNC, D-sub, and/or USB plugs-and-sockets) (\textbf{Fig.~\ref{fig:franka_insertion_env}}). The assets can be randomized in type and location across all environments. The default goal is to pick up a peg or plug and insert it into its corresponding hole or socket.
    \item \textbf{FrankaGearsEnv}, which contains a Franka robot and a $4$-part gear assembly (\textbf{Fig.~\ref{fig:franka_gears_env}}). The assets can be randomized in location across all environments. The default goal is to pick up each gear, insert it onto its corresponding gear shaft, and align it with any other gears. 
\end{itemize}

\subsection{Controllers}
Research efforts in reinforcement learning for robotic manipulation have traditionally used an action space consisting of low-level position, velocity, or torque commands. On the other hand, classical PD- or PID-style robot controllers have been used to solve contact-rich tasks in robotic assembly for several decades \cite{mason_mechanics_2001, whitney_mechanical_2004}. In recent years, there has been substantial interest in using an RL action space consisting of targets to such controllers, with promising results in both sample efficiency and asymptotic performance \cite{martin-martin_variable_2019, peng_learning_2017}. 

Akin to \cite{zhu_robosuite_2020} in MuJoCo, we provide a series of robot controllers based on those that researchers and engineers commonly use in the real world. The actions of the controllers are executed using an explicit integrator to avoid undesired damping. The controllers are as follows:
\begin{itemize}
\item \textbf{Joint-space inverse differential kinematics (IK) motion controller}, which converts task-space errors into joint-space errors and applies PD gains to generate joint torques. The IK controller can use either the geometric or analytic Jacobian \cite{siciliano_robotics_2009} and generate torques with the Jacobian pseudoinverse, Jacobian transpose, damped least-squares (Levenberg-Marquardt), or adaptive SVD \cite{buss_introduction_2009}.
\item \textbf{Joint-space inverse dynamics (ID) controller}, which uses the joint-space inertia matrix and gravity compensation to generate joint torques, achieving desired spring-damper behavior in joint-space \cite{eth_dynamics_2018}.
\item \textbf{Task-space impedance controller}, which applies PD gains to task-space errors to generate joint torques. This controller is immediately available on the real-world Franka robot via the \textit{libfranka} library \cite{franka_libfranka_2017}.
\item \textbf{Operational-space (OSC) motion controller}, which uses the task-space inertia matrix and gravity compensation to generate joint torques, achieving desired spring-damper behavior in task-space (akin to \cite{wong_oscar_2021}).
\item \textbf{Open-loop force controller}, which converts a task-space force target into joint torques.
\item \textbf{Closed-loop P force controller}, which stacks an open-loop force controller with a closed-loop controller that applies P gains to task-space force errors. 
\item \textbf{Hybrid force-motion controller}, which stacks a task-space impedance or OSC motion controller with an open- or closed-loop force controller. Selection matrices can specify which axes use motion and/or force control.
\end{itemize}

Mathematical formulations are provided in \textbf{App.~\ref{sec:app_controllers}}. 

\section{Reinforcement Learning}

The robotics community has demonstrated that RL can effectively solve simulated or real-world assembly tasks. However, these efforts are often limited to off-policy algorithms, require extensive training time or human demonstrations/corrections, and/or only address simple tasks. With our contact simulation methods, we use on-policy RL to solve the most contact-rich task on NIST Task Board $1$: assembling a nut onto a bolt. Like many assembly tasks, such a procedure is long-horizon and challenging to learn end-to-end. We divide the task into $3$ phases and learn an subpolicy for each:
\begin{itemize}
    \item \textbf{Pick}: The robot grasps the nut with a parallel-jaw gripper from a random location on a work surface.
    \item \textbf{Place}: The robot transports the nut to the top of a bolt fixed to the surface.
    \item \textbf{Screw}: The robot brings the nut into contact with the bolt, engages the mating threads, and tightens the nut until it contacts the base of the bolt head.
\end{itemize}

\begin{figure}
    \centering
    \includegraphics[width=\columnwidth]{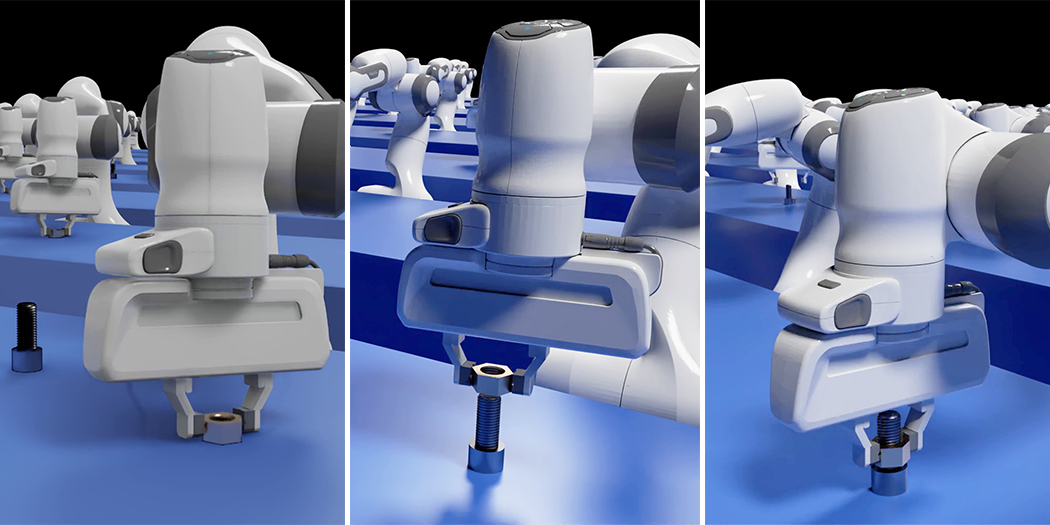}
    \caption{Rendering of achieved goal states of our trained subpolicies for \textbf{FrankaNutBoltEnv}. Left: \textbf{Pick}. Middle: \textbf{Place}. Right: \textbf{Screw}.}
    \label{fig:franka_nut_bolt_rl}
\end{figure}

RL is neither the only means to solve the $3$ phases of this task, nor the most efficient: \textbf{Pick} and \textbf{Place} can be solved with classical grasping and motion controllers, and although challenging, \textbf{Screw} may be solved using a nut-driver and a compliance and/or suction mechanism \cite{kuka_fast_2016}. We investigate this task as a proof-of-concept that our simulation methods can enable efficient policy learning for tasks of such complexity. Moreover, it is a common experience of simulation developers that model-free RL agents reveal and exploit any inaccuracies or instabilities in the simulator to maximize their reward; we view successfully training RL agents in contact-rich tasks as important qualitative evidence of simulator robustness.

We describe each subpolicy below; detailed evaluations will focus on \textbf{Screw}, the most contact-rich of the phases. We then address sequential execution and examine contact forces. 

\subsection{Shared Framework}

The \textbf{Pick}, \textbf{Place}, and \textbf{Screw} subpolicies were all trained in Isaac Gym using our simulation methods and FrankaNutBoltEnv environment. The PPO implementation from \cite{makoviichuk_rl-games} was used with a shared set of hyperparameters (\textbf{Table~\ref{tab:ppo_parameters}}). Typically, a batch of $3\mbox{-}4$ policies were trained simultaneously on a single NVIDIA RTX 3090 GPU, with each policy using $128$ parallel simulation environments. Each batch required a total of $1\mbox{-}1.5$ hours for $1024$ policy updates.

We defined our action space as targets for our implemented controllers. Unless otherwise specified, the targets for the joint-space IK controller, joint-space ID controller, task-space impedance controller, and OSC motion controller were all $6$-DOF transformations \textit{relative} to the current state, with the rotation expressed as axis-angle. The targets for the open-loop and closed-loop force controller were $3$-dimensional force vectors. The targets for the hybrid force-motion controller were the $9$-dimensional union of the previous $2$ action spaces.

We now discuss our randomization, observations, rewards, success criterion, and success rate for each subpolicy.

\subsection{Subpolicy: Pick}

At the start of each \textbf{Pick} episode, the $6$-DOF Franka hand pose and $2$-DOF nut pose (constrained by the work surface) were randomized over a large spatial range (\textbf{Table~\ref{tab:domain_randomization}}). 

\begin{table*}
\centering
\begin{tabular}{ll|ll|ll}
\multicolumn{2}{c|}{Pick} & \multicolumn{2}{c|}{Place} & \multicolumn{2}{c}{Screw} \\ \hline
\textbf{Parameter} & \textbf{Range} & \textbf{Parameter} & \textbf{Range} & \textbf{Parameter} & \textbf{Range} \\ \hline
\rowcolor{gray!10} Hand X-axis & [-0.2, 0.2] m & Hand XY-axes & [-0.2, 0.2] m & Hand angle & [-90, 90{] deg} \\
Hand Y-axis & [-0.4, 0.0] m & Hand Z-axis & [0.5, 0.7] m & Fingertip X-axis & [-3, 3{] mm} \\
\rowcolor{gray!10} Hand Z-axis & [0.5, 0.7] m & Hand roll, pitch & [-17, 17] deg & Fingertip Y-axis & [-3, 3{] mm} \\
Hand roll, pitch & [-17, 17] deg & Hand yaw & [-57, 57] deg & Fingertip Z-axis & [0, 3] mm \\
\rowcolor{gray!10} Hand yaw & [-57, 57] deg & Nut-in-gripper XY-axes & [-2, 2] mm & Nut-in-gripper X-axis & [-3.5, 3.5] mm \\
Nut X-axis & [-0.1, 0.1] m & Nut-in-gripper Z-axis & [-5, 5] mm & Nut-in-gripper Z-axis & [-6.5, 1.0] mm \\
\rowcolor{gray!10} Nut Y-axis & [-0.4, 0.2] m & Nut-in-gripper yaw & [-180, 180] mm & Nut-in-gripper yaw & [-15, 15] deg \\
 &  & Bolt XY-axes & [-10, 10] mm &  & 
\end{tabular}
\caption{\label{tab:domain_randomization}Ranges for initial randomization of pose parameters. Parameter values were uniformly sampled from the ranges.}
\end{table*}

The observation space for \textbf{Pick} was the pose (position and quaternion) of the hand and nut, as well as the linear and angular velocity of the hand. In the real world, the pose and velocity of the hand can be determined to reasonable accuracy ($<1~cm$) through a forward kinematic model and proprioception of joint positions and velocities, whereas the pre-grasp pose of the nut (a known model, as typical in industrial settings) can be accurately estimated through pose estimation frameworks~\cite{labbe_eccv_2020}. The action space for \textbf{Pick} consisted of joint-space IK controller targets with damped least-squares. 

A dense reward was formulated as the distance between the fingertips and the nut. Initial experiments defined this distance as $\mathbf{||x||} + \alpha \mathbf{||q||}$, where $\mathbf{x}$ and $\mathbf{q}$ are the translation and quaternion errors, and $\alpha$ is a scalar hyperparameter. However, this approach was sensitive to $\alpha$. Inspired by \cite{allshire_transferring_2021}, we reformulated the distance as $\mathbf{||k_n - k_f||}$, where $k_n$ and $k_f$ are both tensors of $2\mbox{-}4$ keypoints distributed along the nut central axis and end-effector approach axis, respectively. Intuitively, this method computes distance on a single manifold, obviating tuning. The collinearity of each keypoint set also allows equivariance to rotation of the hand (i.e., yaw) about the nut central axis.

After executing the \textbf{Pick} subpolicy for a prescribed (constant) number of timesteps, a manually-specified grasp-and-lift action was executed. Policy success was defined as whether the nut remained in the grasp after lifting. If successful, a success bonus was added to the episodic return.

With the above approach, the \textbf{Pick} policy was able to achieve a $100\%$ success rate within the randomization bounds. Qualitatively, the agent learned to execute a fast straight-line path towards the nut, followed by a slow pose refinement. Due to the collinearity of the keypoints, the final pose distribution of the hand was highly multimodal in yaw.

\subsection{Subpolicy: Place}

At the start of each \textbf{Place} episode, the Franka hand and nut were reset to a known stable grasp pose. The nut-in-gripper position/rotation and the bolt position were randomized. The hand-and-nut were moved to a random pose using the joint-space IK controller (\textbf{Table~\ref{tab:domain_randomization}}). Training was then initiated.

The observation space for \textbf{Place} was identical to that for \textbf{Pick}, but also included the pose (position and quaternion) of the bolt. When grasped, the nut pose may be challenging to determine in the real world; however, recent research has demonstrated that visuotactile sensing with known object models can enable high-accuracy pose estimates \cite{bauza_tactile_2020}.

The action space was identical to that for \textbf{Pick}. A dense reward was again formulated as a keypoint distance, now between the bolt and nut central axes. The keypoints were defined such that, when perfectly aligned, the base of the nut was located $1~mm$ above the top of the bolt. Success was defined as when the average keypoint distance was $<0.8~mm$

With the above approach, the \textbf{Place} policy was able to achieve a 98.4\% success rate within the randomization bounds. A common initial failure case during training was collision between the gripper and the bolt, dislodging the nut. The robot learned trajectories that remained above the top plane of the bolt, with a slow pose refinement phase when close.

Although having negligible effect on steady-state error, an effective strategy for smoothing the \textbf{Place} trajectory was applying an \textit{action gradient penalty} at each timestep. The penalty was equal to $\beta ||\mathbf{a_t} - \mathbf{a_{t-1}}||$, where $\mathbf{a}$ is the $6$-dimensional action vector and $\beta$ is a hyperparameter ($0.1$).

\subsection{Subpolicy: Screw}\label{sec:screw_policy} At the start of each \textbf{Screw} episode, the Franka hand and nut were reset to a stable grasp pose, randomized relative to the top of the bolt (\textbf{Table~\ref{tab:domain_randomization}}); these stable poses were generated using the \textbf{FrankaCalibrate} script described in \textbf{App.~\ref{sec:app_helpful}}. The nut-in-gripper position was also randomized as before.

Among the subpolicies, \textbf{Screw} was by far the most contact-rich, and as follows, challenging to train. The robot was required to bring the nut into contact with the bolt, engage the respective threads, generate precise torques along the $7$ arm joints to allow the high-inertia robot links to admit the rigid bolt constraint, and maintain appropriate posture of the gripper with respect to the nut during tightening. As a simplifying assumption, the joint limit of the end-effector was removed, allowing the Franka to avoid regrasping (akin to the Kinova Gen3). Nevertheless, training was replete with a diverse range of pathologies, including high-energy collision with the bolt shank, roll-pitch misalignment of the nut when first engaging the bolt threads, jamming of the nut during tightening, and precession of the gripper around the bolt during tightening, which induced slip between the gripper and nut.

To overcome the preceding issues, a systematic exploration of controllers/gains, observation/action spaces, and baseline rewards was executed. First, policies for $3$ task-space controllers were evaluated over a wide range of gains, and the controller-gain configuration with the highest success rate was chosen (\textbf{Table~\ref{tab:controller_evaluation}}). Then, $4$ observation spaces were evaluated, and the space with the highest success rate was selected (\textbf{Table~\ref{tab:observation_evaluation}}). The procedure continued with $2$ action spaces (\textbf{Table~\ref{tab:action_evaluation}}) and $3$ baseline rewards (\textbf{Table~\ref{tab:reward_evaluation}}). Success was defined as when the nut was less than $1$ thread away from the base of the bolt.

To encourage stable robot posture, a dense reward was formulated that consisted of the sum of the keypoint distance [between nut and base of bolt] and [between end-effector and nut]. To prioritize both task completion and efficient training, early termination was applied on success and failure, and a maximum of $1024$ gradient updates was allowed. Future work will investigate asymptotic performance with more updates.

Notably, collisions between the complex geometries of the nut and bolt remained stable during exploration by the RL agent. However, the majority of experimental groups failed due to the pathologies described earlier. The highest performing agents consistently used an OSC motion controller with low proportional gains, an observation space consisting of pose and velocity of the gripper and nut, a $2$-DOF action space ($Z$-translation and yaw), and a linear baseline reward. As expected, the relatively low number of epochs biased towards lower-dimensional observations and actions. 

Using the above configuration, a final \textbf{Screw} policy was trained over $4096$ gradient updates and achieved an $85.6\%$ success rate over $1024$ episodes.

\begin{table*}
\centering
\begin{tabular}{lrcrr}
\textbf{Observations} & \textbf{Success Rate} & \textbf{Env Steps to Success} & \textbf{Reward} & \textbf{Joint Torque (Nm)} \\ \hline
\rowcolor{gray!10} Pose & 0.7708 & \textbf{2318} & -0.1019 & \textbf{1.7319}  \\
Pose, velocity & \textbf{0.7760} & 3015 & -0.0941 & 1.7330 \\
\rowcolor{gray!10} Pose, velocity, force & 0.5026 & 3849 & -0.0784 & \textbf{1.2186} \\
Pose, velocity, force, action & 0.3307 & 3791 & \textbf{-0.0521} & 1.2257
\end{tabular}
\caption{\label{tab:observation_evaluation}Comparison of observation spaces on performance of \textbf{Screw} task. \textit{Success Rate} specifies the fraction of episodes that were successful. \textit{Time to Success} specifies the mean number of timesteps required to achieve success. \textit{Reward} specifies the mean reward during each episode. \textit{Joint Torque} specifies the mean joint torque norm ($Nm$) during each episode. Each cell is computed from the average of $3$ seeds. \textit{Pose} was quickest; \textit{Pose, velocity} exhibited highest success rate; and \textit{Pose, velocity, force, action} achieved lowest mean reward, but did not consistently complete the task.}
\end{table*}

\subsection{Sequential Policy Execution}

Although not our primary focus, a natural question arose on whether the subpolicies could be chained. Policy chaining can be challenging, as errors in each subpolicy can accumulate into poor overall performance; as a simple example, $3$ perfectly-coupled subpolicies with $90\%$ success rates can produce a combined policy with a $72.9\%$ success rate.

In this work, we used a simple strategy to connect the learned \textbf{Pick}, \textbf{Place}, and \textbf{Screw} subpolicies end-to-end. Specifically, when training a given subpolicy, the initial states were randomized to span the distribution of the final states of the preceding trained subpolicy. For example, we defined the initial states of the \textbf{Screw} subpolicy (\textbf{Table~\ref{tab:domain_randomization}}) to span the maximum observed error of the \textbf{Place} subpolicy. For a small number of subpolicies, this strategy may be effective; however, the approach does not scale to long sequences, as Policy $N$ must be trained and Sequence $0...N$ must be evaluated before training Policy $N + 1$. To facilitate smoothness, an exponential moving average was applied on \textbf{Place} actions.

With this strategy, we achieved an end-to-end \textbf{Pick}, \textbf{Place}, and \textbf{Screw} success rate of $74.2\%$. More sophisticated techniques can be explored for improvements \cite{clegg_learning_2018, lee_adversarial_2021}.

\subsection{Contact Forces}

Quantitatively and through numerous visual comparisons, our physics simulation module enabled accurate, efficient, and robust simulation of contact-rich interactions of assets with real-world geometries and material properties. Furthermore, our module was built on  PhysX, which has been evaluated under challenging sim-to-real conditions \cite{allshire_transferring_2021, da_learning_2018, rudin_learning_2021}.

\begin{figure}
    \centering
    \includegraphics[height=2in]{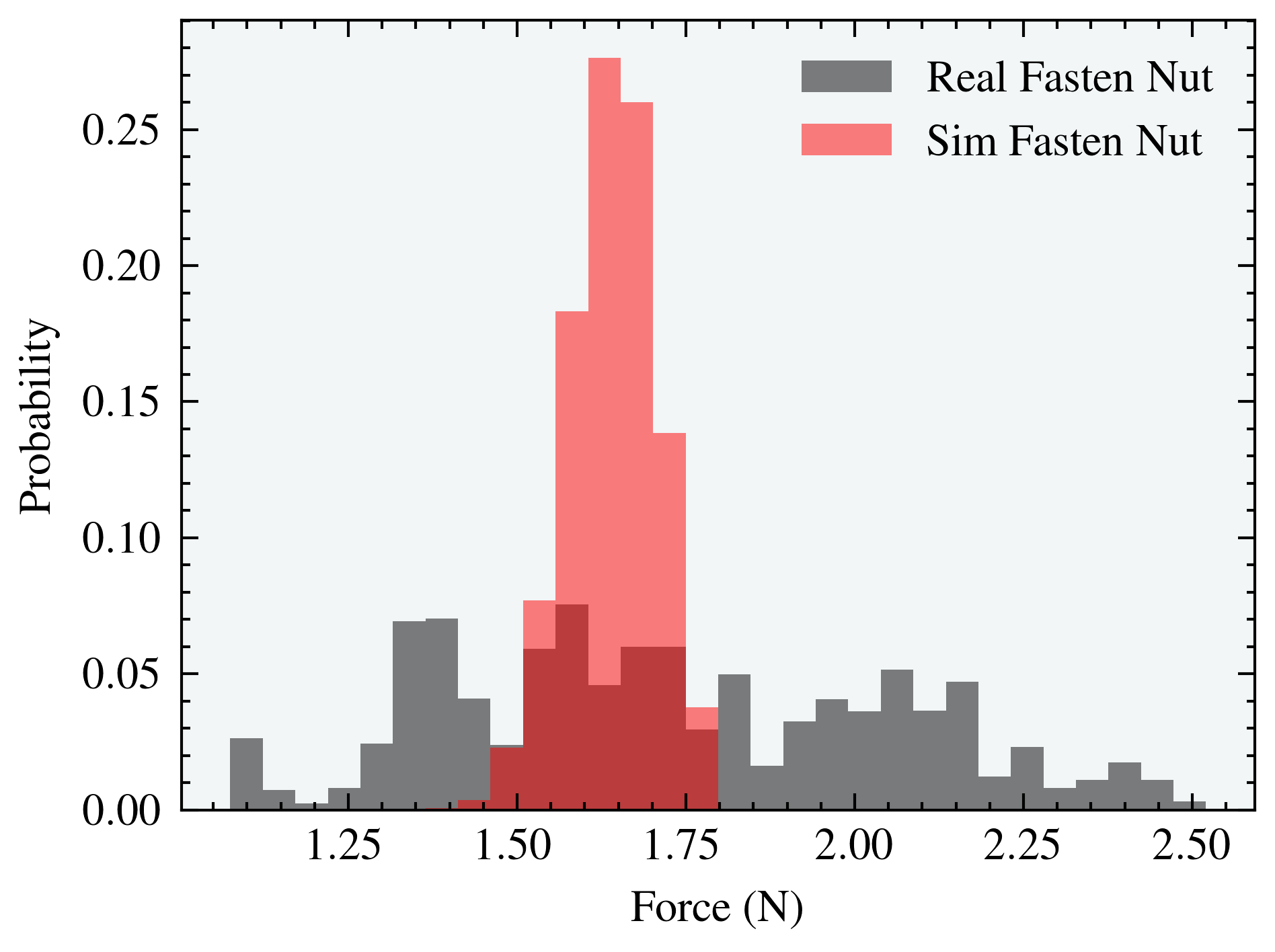}
    \caption{Comparison of simulated contact forces during \textbf{Screw} subpolicy execution with analogous real-world contact forces from the Daily Interactive Manipulation (DIM) dataset \cite{huang_dataset_2018}. In this DIM task, humans tightened nuts using a wrench outfitted with a force-torque sensor. Extreme outliers were rejected for visualization. Maximum mean discrepancy (MMD) was $0.01269$. (MMD values were $\mbox{$\sim$}0.1\mbox{-}1.0$ for less-related tasks.)}
    \label{fig:sim_vs_real_histogram}
\end{figure}

Nevertheless, it is also important to consider the contact forces generated during such interactions. We executed our \textbf{Screw} subpolicy and recorded joint torque norms, as well as contact force norms at the gripper fingers and bolt (\textbf{Fig.~\ref{fig:validation_force_ranges}}). The joint torques are well within the range of lightweight collaborative robots (\textit{e.g.}, UR3). Furthermore, the contact force norms at the fingertips were compared to analogous real-world forces from the Daily Interactive Manipulation dataset \cite{huang_dataset_2018}, in which human subjects tightened or loosened nuts with a wrench outfitted with a force-torque sensor (\textbf{Fig.~\ref{fig:sim_vs_real_histogram}}).

Although the reward functions for the RL agents never involved contact forces, the robots learned policies that generated forces in the middle of human ranges; the much higher variance of human forces was likely due to more diverse strategies adopted by humans. Combined with our visual comparisons, these results do not guarantee sim-to-real policy transfer, but demonstrate that raw quantities computed by simulation are highly comparable to the real world.

\section{Discussion}

We have presented \textbf{Factory}, a set of physics simulation methods and robot learning tools for contact-rich interactions in robotics. We provide a physics simulation module for PhysX and Isaac Gym that enables $100$s to $1000$s of contact-rich interactions to be simulated in real-time on a single GPU, as tested on a diverse array of scenes. As one example, $1000$ nuts-and-bolts were simulated in real-time, whereas the established benchmark was a single nut-and-bolt at $\frac{1}{20}$ real-time. 

We also provide $60$ carefully-designed, ISO-standard or manufacturer-based assets from the NIST Assembly Task Board 1, suitable for high-accuracy simulation; $3$ robotic assembly scenes in Isaac Gym where a robot can interact with these assets across a diverse range of assembly operations (fastener tightening, insertion, gear meshing); and $7$ classical robot controllers that can achieve pose, force, or hybrid targets. We intend for our assets, environments, and controllers to grow over time with contributions from the community.

Finally, we train proof-of-concept RL policies in Isaac Gym for the most contact-rich interaction on the board, nut-and-bolt assembly. We show that we can achieve stable simulator behavior, efficient training ($1\mbox{-}1.5$ hours to simultaneously train $4$ policies on $1$ GPU), high success rates, and realistic forces/torques. Although \textbf{Factory} was developed with robotic assembly as a motivating application, there are no limitations on using our methods for entirely different tasks within robotics, such as grasping of complex non-convex shapes in home environments, locomotion on uneven outdoor terrain, and non-prehensile manipulation of aggregates of objects.

\section{Limitations}

We plan to address several limitations of this work. Within simulation, we plan to make $3$ improvements to our SDF collision scheme: 1) the ability to robustly handle collisions of thin-shell meshes (\textit{e.g.}, thin-walled bottles and boxes), 2) improved handling of low-tessellation meshes, as currently, $1$ contact is generated per-triangle, allowing penetration on large flat surfaces, 3) using sparse SDF representations to reduce the SDF memory footprint (\textbf{Table~\ref{tab:simulation_geometric_representations}}). Furthermore, we are adding support for FEM-based simulation of stiff deformable features, such as the flexible tab on an RJ45 connector.

Within our assets, environments, and controllers, we plan to add assets for additional industrial and home subassemblies (\textit{e.g.}, USB-C, power plugs, key-in-lock), scenes for additional assembly tasks (\textit{e.g.}, chain-and-sprocket assembly), and controllers found in industrial settings (\textit{e.g.}, admittance). Within policy training, we plan to extend our policy for \textbf{FrankaNutBoltEnv} to learn regrasp behavior. In addition, we aim to develop a unified proof-of-concept policy for all insertion tasks within \textbf{FrankaInsertionEnv}, as well as a policy for gear meshing within \textbf{FrankaGearsEnv}, further evaluating training efficiency and simulator robustness. However, we encourage the broader RL community to test and develop state-of-the-art RL algorithms around these complex tasks.

\subsection{Future Work}

Upon making the aforementioned improvements, our future work will focus primarily on sim-to-real transfer. As described earlier, there has been compelling evidence  sim-to-real is possible for industrial insertion tasks; we aim to demonstrate this for more complex bolt-tightening and gear-meshing tasks, as well as full $3D$ assembly operations in both industrial and home settings. For perception, we may train image-based policies using real-time ray-tracing and/or post-simulation path-tracing \cite{blender_2018, nvidia_omniverse_2022} combined with domain randomization \cite{tobin_domain_2017}, However, we find distillation approaches to be particularly compelling. Specifically, we can
\begin{enumerate}
    \item Train RL \textit{teacher} policies, which take privileged information (\textit{e.g.}, $6$-DOF states) as input; and then use imitation learning to learn \textit{student} policies, which take images as input and replicate the teacher's actions \cite{chen_system_2021, lee_learning_2020}, or
    \item Train RL policies with the actor accessing images, but the critic accessing privileged information \cite{andrychowicz_learning_2020}.
\end{enumerate}

Adding noise on both low-dimensional and high-dimensional observations may be valuable. Furthermore, given that camera observations will be occluded during contact, we anticipate that integrating tactile sensing into our real-world system will be exceptionally critical for object-gripper pose estimation and slip detection. 

Algorithmically, we are interested in obviating human demonstrations or corrections during policy learning. Nevertheless, our physics simulation module is also suitable for simulation-based demonstration collection for use in imitation learning \cite{clever_assistive_2021, mandlekar_roboturk_2018} or DDPGfD-style policy learning.

\section{Conclusion} 
\label{sec:conclusion}

We aim for \textbf{Factory} to establish the state-of-the-art in contact-rich simulation, as well as serve as an existence proof that highly efficient simulation and learning of contact-rich interactions is possible in robotics. Our experience has shown that high-quality assets and accurate, efficient simulation methods drastically reduce the inductive bias and algorithmic burden required to solve contact-rich tasks. We invite the community to establish benchmarks for solving the provided scenes, as well as extend and use \textbf{Factory} for their own contact-rich applications both within and outside of RL. We also hope that this work inspires researchers to execute contact-rich simulations of tasks beyond what we show in this paper to enable further solutions to complex problems.

\section*{Author Contributions}
YN led the research project.

MM initially developed SDF collisions for FleX.

MM and YN conducted proof-of-concept demonstrations of SDF collisions for robotic assembly in FleX.

KS, ML, PR, and AM developed SDF collisions and contact reduction for PhysX.

PR, KS, AM, and YN developed evaluation and demonstration scenes for PhysX.

LW, YG, and GS integrated the PhysX developments into Isaac Gym.

YN, IA, and PR developed the assets, environments, and controllers for Isaac Gym.

YN, IA, and AH developed the RL policies within Isaac Gym.

DF, AH, MM, ML, GS, and AM advised the project.

YN, MM, AH, and IA wrote the paper.

\section*{Acknowledgments}
We thank Joe Falco for assistance with the original NIST assets, John Ratcliff for generating the convex decomposition for the bolt (\textbf{Fig.~\ref{fig:convex_decomposition}}), Karl Van Wyk and Lucas Manuelli for helpful discussions on controllers, Viktor Makoviychuk for assistance with the \textit{rl-games}~\cite{makoviichuk_rl-games} library, and the RSS reviewers for their insightful comments and questions.

\bibliographystyle{plainnat}
\bibliography{references}

\appendix
\renewcommand\thefigure{S\arabic{figure}}

\subsection{Introduction}
No supplementary information.

\subsection{Related Works}
No supplementary information.

\subsection{Contact-Rich Simulation Methods}

\subsubsection{Initial Nut-and-Bolt Experiments\label{sec:app_init_nut_bolt_exp}}

\begin{figure}
    \centering
    \includegraphics[width=\columnwidth]{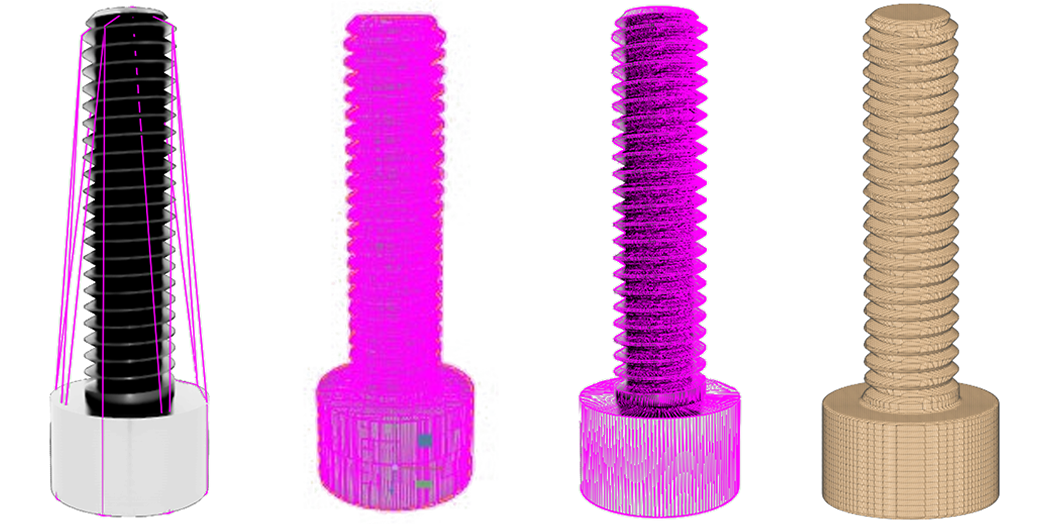}
    \caption{Geometric representations, demonstrated on an M4 bolt. Left to right: Convex hull. Convex decomposition generated via V-HACD \cite{mamou_volumetric_2016} ($951$ shapes). Triangle mesh generated in Onshape ($47k$ triangles). SDF ($144 \times 256 \times 144$ voxels), visualized as a mesh of the isosurface.}
    \label{fig:geometric_representations}
\end{figure}

We performed initial experiments with various geometric representations of nuts and bolts before proceeding with SDFs. \textbf{Fig.~\ref{fig:geometric_representations}} shows 4 different geometric representations of an M4 bolt mesh.

\begin{figure}
    \centering
    \includegraphics[width=\columnwidth]{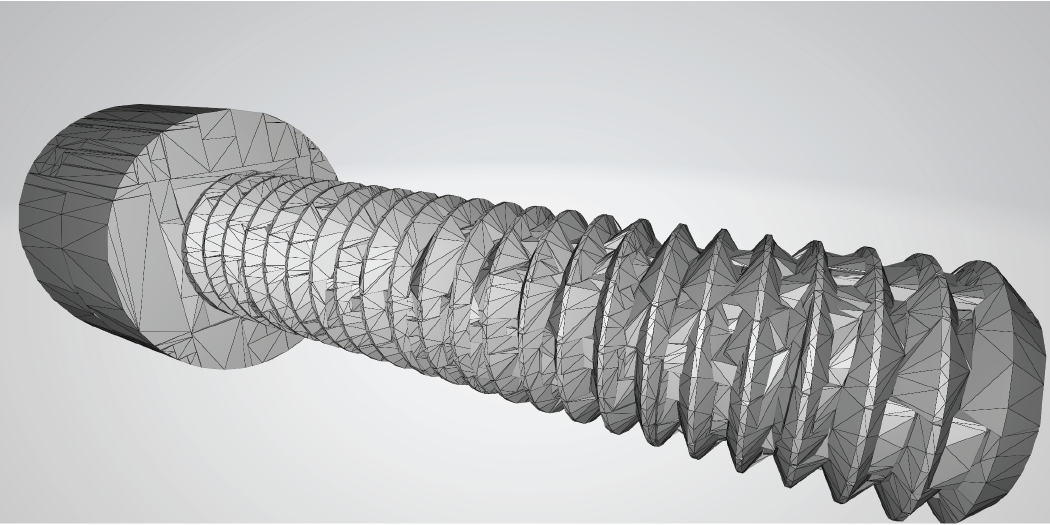}
    \caption{Automatic convex decomposition from \textbf{Fig.~\ref{fig:geometric_representations_supp}} for an M4 bolt ($951$ shapes). Spatial artifacts are apparent.}
    \label{fig:convex_decomposition}
\end{figure}

The convex representation is clearly insufficient to simulate contact-rich interactions. The convex decomposition, generated with voxel-based V-HACD using Omniverse \cite{nvidia_omniverse_2022}, consists of $951$ convex hulls and appears to faithfully represent the exact bolt geometry. However, \textbf{Fig.~\ref{fig:convex_decomposition}} shows a closer view; artifacts are prevalent. These artifacts cause inaccuracy and instability when simulating contact with a mating M4 nut.

\begin{figure}
    \centering
    \includegraphics[width=\columnwidth]{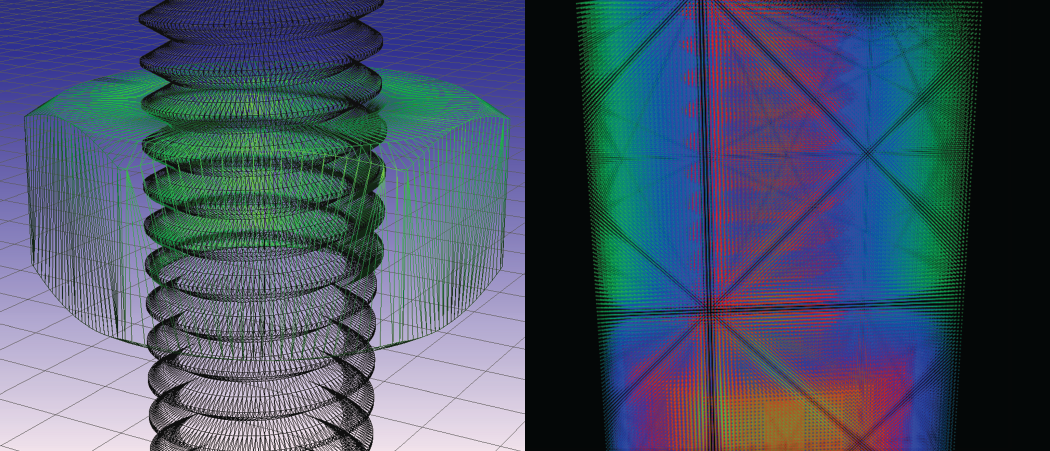}
    \caption{Left: High-quality triangular mesh representation of an M4 nut and bolt, consisting of $27k$ triangles (nut) and $47k$ triangles (bolt). Right: SDF representation of the bolt, stored in a voxel grid of dimensions $144 \times 256 \times 144$ and visualized as a point cloud. For easier visualization, colors are interpolated according to ${\phi(\boldsymbol{x})}^{\frac{1}{4}}$. Streaks are artifacts of the viewing angle.}
    \label{fig:geometric_representations_supp}
\end{figure}

The trimesh representation, consisting of $47k$ triangles, is highly faithful to the exact geometry. \textbf{Fig.~\ref{fig:geometric_representations_supp}} shows the M4 bolt and nut meshes. Initial experiments were conducted with trimesh-trimesh collisions using boundary-layer expanded meshes \cite{hauser_robust_2016}. However, stable behavior was only possible with extremely small timesteps ($\Delta t = 1e\mbox{-}3$ s with $10$ substeps), and the resulting simulations executed at $\frac{1}{80}$ of real-time.

Only SDF representations enabled accurate, efficient, and robust simulation. \textbf{Fig~\ref{fig:geometric_representations_supp}} shows a point-cloud visualization of the SDF for the M4 bolt mesh. However, generating SDFs is an expensive process proportional to the number of samples, and storage requires significant memory. The M4 bolt model consists of $5.3$ million samples, each a floating-point distance.

\begin{table*}[b]
\centering
\begin{tabular}{llllll}
 & \multicolumn{2}{c}{Stable Timesteps} & \multicolumn{2}{c}{Memory Bandwidth} & \multicolumn{1}{c}{Parallelization} \\ \hline
\textbf{Solver type} & \textbf{Substeps} & \textbf{Iterations} & \textbf{Per frame} & \textbf{Per second} & \textbf{Max nuts/bolts} \\ \hline
\rowcolor{gray!10} Jacobi (before) & 8 & 64 & 1.28 GB & 76.8 GB & 20 \\ 
Jacobi (after) & 8 & 64 & 24 MB & 1.44 GB & 1100 \\ 
\rowcolor{gray!10} Gauss Seidel (before) & 1 & 16 & 40 MB & 2.4 GB & 666 \\ 
Gauss Seidel (after) & 1 & 16 & 768 KB & 46 MB & 35000 \\ 
\end{tabular}
\caption{\label{tab:solver_comparison}Comparison of memory bandwidth requirements for a Jacobi and Gauss-Seidel solver on the nut-and-bolt scene, before and after contact reduction. \textit{Max nuts/bolt} denotes the theoretical maximum number of nuts and bolts that could be simulated on a state-of-the-art GPU; in practice, limitations aside from contact constraint solution (\textit{e.g.}, scaling of SDF-based contact generation) typically reduce this upper bound by one order of magnitude. Note that although \textit{Gauss-Seidel (before)} has lower memory bandwidth requirements than \textit{Jacobi (before)}, it is not preferred due to its low execution speed.}
\end{table*}

\begin{figure}
    \centering
    \includegraphics[width=\columnwidth]{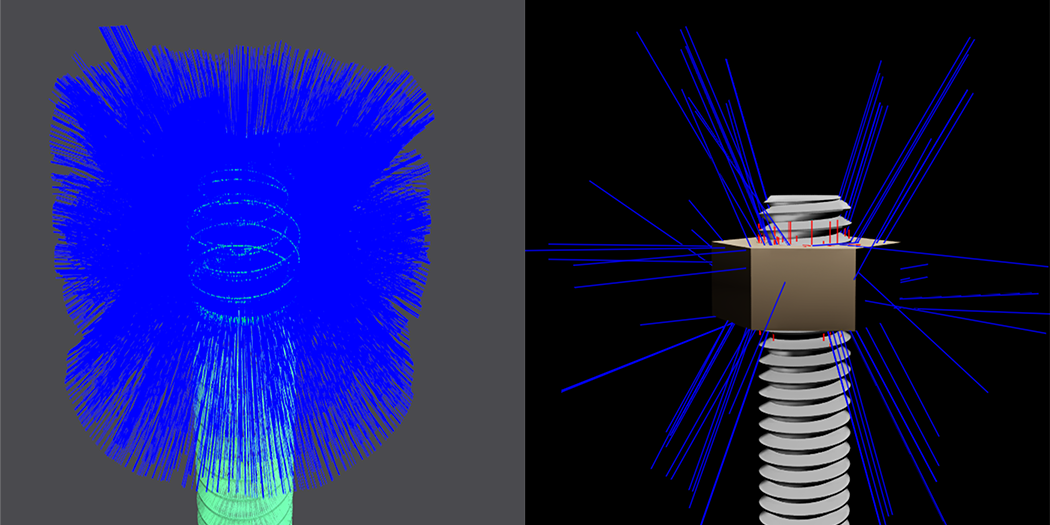}
    \caption{Contacts generated between a nut and bolt before and after applying our contact reduction scheme. Left: $16k$ contacts generated on the bolt before reduction. Right: $300$ contacts remaining after reduction.}
    \label{fig:contact_reduction}
\end{figure}

\begin{figure}
    \centering
    \includegraphics[width=\columnwidth]{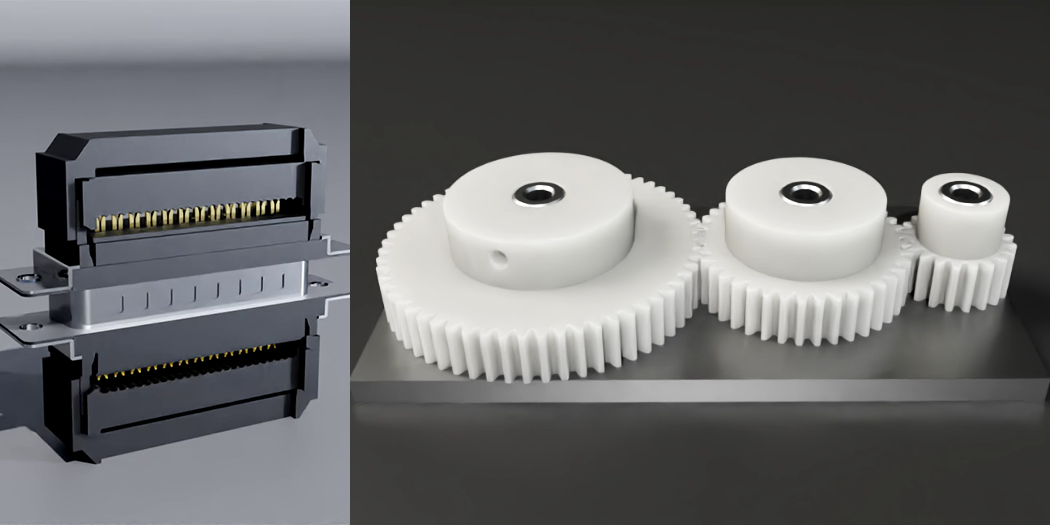}
    \caption{Renderings of D-sub connector and gear assembly assets used in simulation evaluations and provided in this work.}
    \label{fig:nist_dsub_gears}
\end{figure}

\subsubsection{Additional Scene Descriptions}
\label{sec:app_additional_scenes}

\textbf{Franka robot + M16 nut-and-bolt assemblies}. A vibratory feeder is an ubiquitous mechanism for conveying and isolating mechanical components. The feeder consists of a tray or bowl that vibrates an aggregate of parts at high-frequency and small-amplitude. Under gravity, the parts gradually move towards a singulation mechanism, which isolates them for inspection or downstream handling. We demonstrate a tray-style inclined feeder that vibrates an aggregate of nuts at $60~Hz$. The nuts towards a channel that only allows one nut to pass at a time. A Franka robot with a hand-scripted controller then grasps a nut from the channel opening and tightens it onto a bolt (\textbf{Fig.~\ref{fig:vibratory_scene}}).

\begin{figure}
    \centering
    \includegraphics[width=\columnwidth]{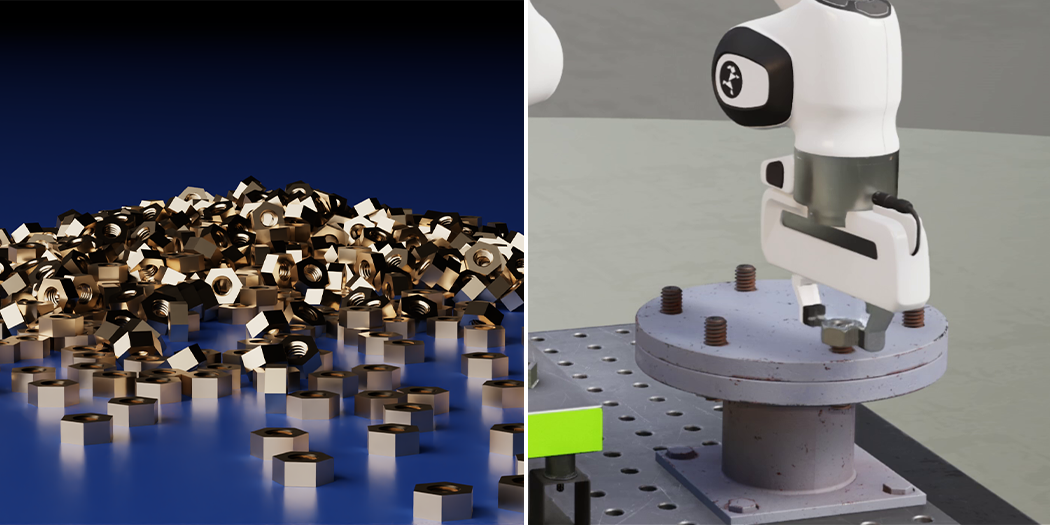}
    \caption{Renderings of 2 scenes. Left: \textbf{M16 nuts} scene. Right: \textbf{Franka robot + M16 nuts + flange assembly scene.}}
    \label{fig:demos}
\end{figure}

\textbf{Franka robot + M16 nuts + flange assembly}. This scene is similar to the above, except that the Franka grasps and tightens multiple nuts in sequence to clamp together $2$ halves of a flange (\textbf{Fig.~\ref{fig:demos}}). This scene is only qualitatively evaluated.

\begin{table*}[b]
\centering
\begin{tabular}{llllllllll}
 & \multicolumn{4}{c}{Representation Info} & \multicolumn{2}{c}{SDF Stats} & \multicolumn{2}{c}{Mesh Stats} \\ \hline
\textbf{Scene} & \textbf{Envs} & \textbf{Bodies} & \textbf{SDF} & \textbf{Mesh} & \textbf{Resolution} & \textbf{Memory} & \textbf{Vertices} & \textbf{Triangles} \\ \hline
\rowcolor{gray!10} Peg-in-hole & 1024 & 2 & Peg & Hole & 43x43x104 & 3.55 MB & 296 & 592 \\
Nut-and-bolt & 1024 & 2 & Bolt & Nut & 144x404x144 & 31.95 MB & 8899 & 17798 \\ 
\rowcolor{gray!10} D-sub connector & 1024 & 2 & Socket & Plug &  534x129x254 & 66.75 MB & 11935 & 23866 \\
Gear assembly & 1024 & 4 & Gear & Gear & 314x314x104 & 39.1 MB & 17557 & 35313 \\
\rowcolor{gray!10} Nuts & 1 & 1024 & Nut & Nut & 284x132x327 & 46.76 MB & 8899 & 17798  \\
Bowls & 1 & 1024 & Bowl & Bowl & 257x75x257 & 14.36 MB & 582 & 1160  \\
\rowcolor{gray!10} Toruses & 1 & 1024 & Torus & Torus & 304x104x304 & 36.60 MB & 1024 & 6144 \\
Franka + nut-and-bolt & 128 & 44 & Bolt & Nut & 144x404x144 & 31.95 MB & 8899 & 17798 \\
\end{tabular}
\caption{\label{tab:simulation_geometric_representations}Geometric representations used in the simulator test scenes. \textit{Bodies} denotes the number of bodies in each parallel environment. For \textit{Nuts}, \textit{Bowls}, and \textit{Toruses}, all bodies are simulated in the same environment. \textit{SDF} and \textit{Mesh} denote which body in the contact pair is represented as an SDF, and which is represented as a mesh that is queried against the SDF. For \textit{Gear assembly}, statistics are provided for SDF-mesh collisions between the large and medium gears. For \textit{Franka + nut-and-bolt}, which contains more than $2$ types of bodies, the Franka-nut collisions are handled with the Franka gripper fingers as convexes and the nut as a mesh, and the nut-nut collisions are handled with the nuts as convexes.}
\end{table*}

\subsubsection{Rendering}\label{sec:app_rendering}

For visualization purposes, we render several figures using Omniverse. The details are as follows:
\begin{itemize}
    \item The output of the simulation in \textbf{Fig.~\ref{fig:teaser}} was path-traced.
    \item The simulation in \textbf{Fig.~\ref{fig:nuts-and-bolts}} was rendered in real-time.
    \item The scene in \textbf{Fig.~\ref{fig:nist_board_sim}} was path-traced.
    \item The scenes in \textbf{Fig.~\ref{fig:franka_nut_bolt_rl}} were path-traced.
    \item The scenes in \textbf{Fig.~\ref{fig:nist_dsub_gears}} were path-traced.
    \item The scene in \textbf{Fig.~\ref{fig:demos}} was path-traced.
\end{itemize}
For simulation-related figures not listed above, the images were captured directly from PhysX, Isaac Gym, or Onshape.

\subsection{Robot Learning Tools}

\subsubsection{Assets}

\textbf{Table~\ref{tab:assets_standard}} and \textbf{Table~\ref{tab:assets_connectors}} provide details on all assets provided in this work. We intend for this to be a growing database, with contributions from the community.

\begin{table}[b]
\begin{tabular}{llll}
\textbf{Part} & \textbf{Standards} & \textbf{Configurations} & \textbf{Clearance (mm)} \\ \hline
\rowcolor{gray!10} 4 mm round peg & ISO 286 & loose, tight & 0.104-0.112 \\
\rowcolor{gray!10} 4 mm round hole &  &  &  \\
8 mm round peg & ISO 286 & loose, tight & 0.105-0.114 \\
8 mm round hole &  &  &  \\
\rowcolor{gray!10} 12 mm round peg & ISO 286 & loose, tight & 0.206-0.217 \\
\rowcolor{gray!10} 12 mm round hole &  &  &  \\
16 mm round peg & ISO 286 & loose, tight & 0.506-0.517 \\
16 mm round hole &  &  &  \\
\rowcolor{gray!10} 4 mm rect. peg & ISO 286 & loose, tight & 0.11-0.14 \\
\rowcolor{gray!10} 4 mm rect. hole &  &  &  \\
8 mm rect. peg & ISO 286 & loose, tight & 0.134-0.224 \\
8 mm rect. hole &  &  &  \\
\rowcolor{gray!10} 12 mm rect. peg & ISO 286 & loose, tight & 0.1374-0.2274 \\
\rowcolor{gray!10} 12 mm rect. hole &  &  &  \\
16 mm rect. peg & ISO 286 & loose, tight & 0.1576-0.2612 \\
16 mm rect. hole &  &  &  \\
\rowcolor{gray!10} M4 nut & ISO 724, 965 & loose, tight & 0.416-0.736 \\
\rowcolor{gray!10} M4 bolt & ISO 724, 965 & loose, tight &  \\
M8 nut & ISO 724, 965 & loose, tight & 0.848-1.325 \\
M8 bolt & ISO 724, 965 & loose, tight &  \\
\rowcolor{gray!10} M12 nut & ISO 724, 965 & loose, tight & 1.26-1.86 \\
\rowcolor{gray!10} M12 bolt & ISO 724, 965 & loose, tight &  \\
M16 nut & ISO 724, 965 & loose, tight & 1.472-2.127 \\
M16 bolt & ISO 724, 965 & loose, tight &  \\
\rowcolor{gray!10} M20 nut & ISO 724, 965 & loose, tight & 1.879-2.664 \\
\rowcolor{gray!10} M20 bolt & ISO 724, 965 & loose, tight &  \\
Gear base, shafts & ISO 286 & loose, tight & 0.005-0.014 \\
Gear small &  &  &  \\
Gear medium &  &  &  \\
Gear large &  &  & 
\end{tabular}
\caption{\label{tab:assets_standard}Information on peg, hole, nut, bolt, and gear models ($49$ parts, $14$ distinct subassemblies) provided and simulated in this work. \textit{Loose} and \textit{Tight} configurations correspond to the ends of the manufacturing tolerance band specified by the listed standard. \textit{Clearance} designates the corresponding range of clearances between the given part and its mating component (\textit{e.g.}, between peg and hole). Clearance values are $2$-sided (\textit{e.g.}, for a round peg, the clearance is diametral).}
\end{table}

\subsubsection{Using Contact Simulation Methods}

To use the contact simulation methods from this work in Isaac Gym, the user simply needs to add an $\mbox{$<$}sdf\mbox{$>$}$ tag to the $\mbox{$<$}collision\mbox{$>$}$ block of the object that should use SDF collisions (\textbf{Listing~\ref{lst:nut_urdf}}). PhysX will then interpret the $\mbox{$<$}sdf\mbox{$>$}$ tag as follows:

Consider 2 colliding objects, Object A and Object B.
\begin{itemize}
    \item If A and B both have an $\mbox{$<$}sdf\mbox{$>$}$ tag, SDF-mesh collision will be applied. The object with the larger number of features (\textit{i.e.}, triangles) will be represented as an SDF, and the trimesh of the other object will be queried against the SDF to check for collisions and generate contacts.
    \item If A has an $\mbox{$<$}sdf\mbox{$>$}$ tag and B does not, convex-mesh collision will be applied. Object A will be represented as a trimesh, and object B will be represented as a convex.
    \item If neither A nor B has an $\mbox{$<$}sdf\mbox{$>$}$ tag, a default convex-convex collision will be applied.
\end{itemize}

\begin{lstlisting}[language=XML, caption={Example URDF file for a nut asset with SDF collision enabled. The \textit{resolution} field specifies the desired number of voxels along the longest dimension of the object. We have typically used $256$ or $512$.}, label=lst:nut_urdf, captionpos=b]
<?xml version="1.0"?>
<robot name="nut_m16">
    <link name="nut_m16">
        <visual>
            <geometry>
                <mesh filename="nut_m16_tight.obj"/>
            </geometry>
        </visual>
        <collision>
            <geometry>
                <mesh filename="nut_m16_tight.obj"/>
            </geometry>
            <sdf resolution="256"/>
        </collision>
    </link>
</robot>
\end{lstlisting}

\subsubsection{Helpful Additions}\label{sec:app_helpful}

For all assembly assets provided (\textit{e.g.}, nut and bolt), the zero-positions of the parts correspond to the configuration in which they are fully assembled; thus, specifying a goal state for RL is extremely simple for the user.

Furthermore, we provide a helper script called \textbf{FrankaCalibrate}. When training RL policies, it is common to randomize the joint state of the robot at the beginning of each episode in order to facilitate robustness. Simultaneously, if a robot is learning a complex multi-step assembly procedure, it can be helpful to learn short-horizon sub-policies that begin when an object is already grasped. Nevertheless, in practice, it is challenging to randomize the state of a robot \textit{and} a grasped object \textit{while maintaining grasp stability}. Using \textbf{FrankaCalibrate}, the Franka can grasp an object, go to randomized poses, and generate a large set of stable robot-and-object poses that can be imported when subsequently training policies.

\begin{table}[b]
\begin{tabular}{llll}
\textbf{Part} & \textbf{Manufacturer} & \textbf{Configurations} & \textbf{Clearance (mm)} \\ \hline
\rowcolor{gray!10} BNC plug (inner) & Amphenol & visual, collision & 0.1778 \\
\rowcolor{gray!10} BNC plug (outer) & Amphenol & visual, collision &  \\
\rowcolor{gray!10} BNC socket & Molex & visual, collision &  \\
D-sub plug & Assmann & visual, collision & 0.0 \\
D-sub socket & Assmann & visual, collision &  \\
\rowcolor{gray!10} RJ45 plug & Harting & visual, collision & 0.34177 \\
\rowcolor{gray!10} RJ45 socket & Amphenol & visual, collision &  \\
USB plug & Bulgin & visual, collision & 0.0 \\
USB socket & Amphenol & visual, collision &  \\
\rowcolor{gray!10} Waterproof plug & Harting & visual, collision & 0.4 \\
\rowcolor{gray!10} Waterproof socket & Harting & visual, collision & 
\end{tabular}
\caption{\label{tab:assets_connectors}Information on electrical connector assets ($11$ parts, $5$ distinct subassemblies) provided in this work. The rigid subassemblies (BNC, D-sub, USB) are also simulated; although the deformable parts (RJ45 plug, Waterproof socket) can be simulated using PhysX or Isaac Gym, they are not tested here. The BNC plug contains both an inner and outer component coupled by a linear spring. \textit{Manufacturer} simply designates the source of the original CAD model used to generate the visual meshes and design the collision meshes; regardless of the manufacturer, the mating features (\textit{e.g.}, pins, holes) for all parts of a certain type are highly standardized. \textit{Clearance} denotes the minimum clearance between the given part and its mating component (\textit{i.e.}, between plug and socket) at the beginning of insertion. Clearance values are 2-sided (\textit{e.g.}, for the D-sub connector, the clearance is the diametral clearance between the pins and holes). The D-sub and USB connectors begin with zero-clearance, whereas the remaining components achieve zero-clearance during insertion. Links to part-specific webpages are available on our website.}
\end{table}

\begin{figure}
    \centering
    \includegraphics[width=\columnwidth]{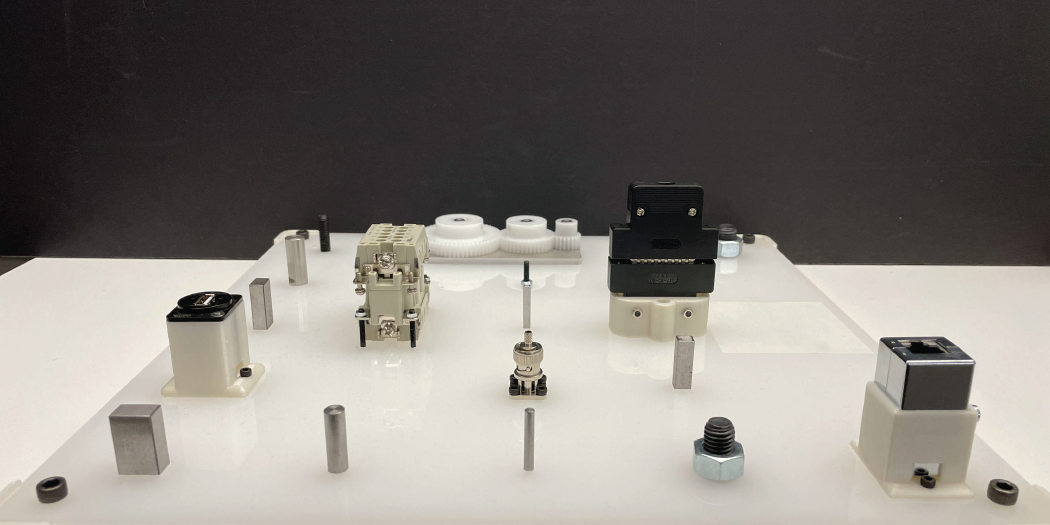}
    \caption{Image of a real NIST Task Board $1$.}
    \label{fig:nist_board_real}
\end{figure}

\subsubsection{Controllers}\label{sec:app_controllers}

The controllers implemented in this work are based on the mathematical formulations articulated in \cite{buss_introduction_2009, kurfess_impedance_2004, lynch_modern_2017, eth_dynamics_2018, siciliano_robotics_2009}. Our specific control laws are as follows:

\begin{itemize}
    \item \textbf{Joint-space inverse differential kinematics (IK) motion controller}
    \begin{equation}
      \boldsymbol{\tau} = \mathbf{k_p} (\mathbf{q_t} - \mathbf{q_c}) - \mathbf{k_d} (\mathbf{\dot q_c})
    \end{equation}
    where $\boldsymbol{\tau}$ is the joint torque vector ($\mathbb{R}^7$), $\mathbf{k_p}$ and $\mathbf{k_d}$ are diagonal matrices consisting of joint-space proportional gains ($\mathbb{R}^{7 \times 7}$) and joint-space derivative gains ($\mathbb{R}^{7 \times 7}$), $\mathbf{q}$ and $\mathbf{\dot{q}}$ are the joint position vector ($\mathbb{R}^7$) and joint velocity vector ($\mathbb{R}^7$), and subscripts $\mathbf{t}$ and $\mathbf{c}$ denote \textit{target} and \textit{current}. The quantity $\mathbf{q_t - q_c}$ is computed by inputting a task-space error into an IK algorithm. For IK, we have implemented Jacobian pseudoinverse, Jacobian transpose, damped least-squares (Levenberg-Marquardt), and adaptive singular value decomposition (SVD) \cite{buss_introduction_2009}.
    
    \item \textbf{Joint-space inverse dynamics controller}
    \begin{equation}
      \boldsymbol{\tau} = \mathbf{M} (\mathbf{k_p} (\mathbf{q_t} - \mathbf{q_c}) - \mathbf{k_d} (\mathbf{\dot q_c})) + \mathbf{b} + \mathbf{g}    
    \end{equation}
    where $M$ is the joint-space inertia matrix ($\mathbb{R}^{7 \times 7}$), $\mathbf{b}$ is the joint-space Coriolis and centrifugal torque vector ($\mathbb{R}^7$), and $\mathbf{g}$ is the joint-space gravitational torque vector (\textit{i.e.}, a gravity compensation term) ($\mathbb{R}^7$). Vector $\mathbf{b}$ is currently ignored due to small joint velocities during assembly.
    
    \item \textbf{Task-space impedance controller}
    \begin{equation}
      \boldsymbol{\tau} = \mathbf{J^T} (\mathbf{k_p} (\mathbf{x_t} \ominus \mathbf{x_c}) - \mathbf{k_d} (\mathbf{\dot x_c}))
    \end{equation}
    where $\mathbf{J}$ is the geometric or analytic Jacobian ($\mathbb{R}^{6 \times 7}$), $\mathbf{k_p}$ and $\mathbf{k_d}$ are now diagonal matrices consisting of task-space proportional gains ($\mathbb{R}^{6 \times 6}$), and $\mathbf{x}$ and $\mathbf{\dot{x}}$ are the task-space position vector ($\mathbb{R}^{6}$) and task-space velocity vector ($\mathbb{R}^{6}$). The analytic Jacobian can be computed from the geometric Jacobian as described in \cite{eth_dynamics_2018}. When using the geometric Jacobian, the orientation components of $\mathbf{\dot{x}}$ are angular velocity, and $\ominus$ computes the rotational transformation from current to target. When using the analytic Jacobian, the orientation components of $\mathbf{\dot{x}}$ are the derivatives of axis-angle, and $\ominus$ simply computes the difference between target and current.
    
    \item \textbf{Operational-space motion controller}
    \begin{equation}
        \boldsymbol{\tau} = \mathbf{J^T} (\mathbf{M} (\mathbf{k_p} (\mathbf{x_t} - \mathbf{x_c}) - \mathbf{k_d} (\mathbf{\dot x_c})) + \mathbf{b} + \mathbf{g})
    \end{equation}
    where $M$ is now the task-space inertia matrix ($\mathbb{R}^{6 \times 6}$), $\mathbf{b}$ is now the task-space Coriolis and centrifugal torque vector ($\mathbb{R}^6$), and $\mathbf{g}$ is now the task-space gravitational torque vector ($\mathbb{R}^6$). The task-space inertia matrix can be computed from the joint-space inertia matrix \cite{khatib_unified_1987}. As before, vector $\mathbf{b}$ is currently ignored due to small velocities during assembly operations.
    
    \item \textbf{Open-loop force controller}
    \begin{equation}
        \boldsymbol{\tau} = \mathbf{J^T} \mathbf{F_t}
    \end{equation}
    where $\mathbf{F_t}$ is the target wrench ($\mathbb{R}^6$).
    
    \item \textbf{Closed-loop P force controller}
    \begin{equation}
        \boldsymbol{\tau} = \mathbf{J^T}(\mathbf{F_t} + \mathbf{k_f} \mathbf{(F_t - F_c)})
    \end{equation}
    where $\mathbf{k_f}$ is a diagonal matrix consisting of task-space proportional gains ($\mathbb{R}^{6 \times 6}$).
    
    \item \textbf{Hybrid force-motion controller}
    \begin{multline}
    \boldsymbol{\tau} = \mathbf{J^T} (\mathbf{S_m} [\mathbf{M} (\mathbf{k_p} (\mathbf{x_t} - \mathbf{x_c}) - \mathbf{k_d} (\mathbf{\dot x_c})) + \mathbf{b} + \mathbf{g}] \\ +  \mathbf{S_f}[\mathbf{F_t} + \mathbf{k_f} \mathbf{(F_t - F_c)}])
    \end{multline}
    where $\mathbf{S_m}$ and $\mathbf{S_f}$ are Boolean diagonal selection matrices that specify which axes use motion control and which use force control ($\mathbb{R}^{6 \times 6}$). Although common practice, the axes need not be complementary.
\end{itemize}

\subsection{Reinforcement Learning}

\textbf{Table~\ref{tab:ppo_parameters}} shows PPO parameters for \textbf{Pick}, \textbf{Place}, and \textbf{Screw}. \textbf{Table~\ref{tab:controller_evaluation}} shows a comparison of controller selection and gains on the \textbf{Screw} task. \textbf{Table~\ref{tab:action_evaluation}} shows a comparison of action spaces on the \textbf{Screw} task. \textbf{Table~\ref{tab:reward_evaluation}} shows a comparison of baseline rewards on the \textbf{Screw} task.

\begin{table}[b]
\centering
\begin{tabular}{ll}
\textbf{Parameter} & \textbf{Value} \\ \hline
\rowcolor{gray!10} MLP network size & [256, 128, 64] \\
Horizon length (T) & 32 \\
\rowcolor{gray!10} Adam learning rate & 1.0e-4 \\
Discount factor ($\gamma$) & 0.99 \\
\rowcolor{gray!10} GAE parameter ($\lambda$) & 0.95 \\
Entropy coefficient & 0.0 \\
\rowcolor{gray!10} Critic coefficient & 2.0 \\
Minibatch size & 512 \\
\rowcolor{gray!10} Minibatch epochs & 8 \\
Clipping parameter $\epsilon$ & 0.2 \\
\end{tabular}
\caption{\label{tab:ppo_parameters} PPO parameters used in pick, place, and screw subpolicies. The actor and critic networks are MLPs with a shared trunk. Observations, actions, advantages, and values were all normalized. The learning rate is small and fixed, as higher returns were consistently achieved compared to an adaptive schedule based on a KL-divergence threshold.}
\end{table}

\begin{table*}[b]
\centering
\begin{tabular}{lrrrrrrrr}
\textbf{Controller} & \multicolumn{1}{l}{\textbf{P Motion Gains}} & \multicolumn{1}{l}{\textbf{D Motion Gains}} & \multicolumn{1}{l}{\textbf{P Force Gains}} & \textbf{Success Rate} & \textbf{Time to Success} & \textbf{Reward} & \textbf{Joint Torque (Nm)} \\ \hline
\multirow{3}{*}{Task-space impedance} & \cellcolor{gray!10} 0.01 & \cellcolor{gray!10} 0.1 & \cellcolor{gray!10} N/A & \cellcolor{gray!10} 0.0 & \cellcolor{gray!10} N/A & \cellcolor{gray!10} -0.321 & \cellcolor{gray!10} 1.993  \\
 & 0.1 & 0.1 & N/A & 0.0 & N/A & -0.308 & 1.966  \\
 & \cellcolor{gray!10} 1 & \cellcolor{gray!10} 0.1 & \cellcolor{gray!10} N/A & \cellcolor{gray!10} 0.0 & \cellcolor{gray!10} N/A & \cellcolor{gray!10} -0.299 & \cellcolor{gray!10} 1.966 \\ \hline
\multirow{3}{*}{OSC motion} & 0.1 & 1 & N/A & \textbf{0.797} & \textbf{2335} & \textbf{-0.101} & \textbf{1.743} \\
 & \cellcolor{gray!10} 1 & \cellcolor{gray!10} 1 & \cellcolor{gray!10} N/A & \cellcolor{gray!10} 0.669 & \cellcolor{gray!10} 2449 & \cellcolor{gray!10} -0.115 & \cellcolor{gray!10} 1.758 \\
 & 10 & 1 & N/A & 0.0 & N/A & -0.305 & 2.891 \\ \hline
\multirow{3}{*}{Hybrid force-motion} & \cellcolor{gray!10} 0.1 & \cellcolor{gray!10} 1 & \cellcolor{gray!10} 0.01 & \cellcolor{gray!10} 0.0 & \cellcolor{gray!10} N/A & \cellcolor{gray!10} -0.411 &  \cellcolor{gray!10} 4.214  \\
 & 1 & 1 & 0.1 & 0.0 & N/A & -1.399 & 28.179 \\
 & \cellcolor{gray!10} 10 & \cellcolor{gray!10} 1 & \cellcolor{gray!10} 1 & \cellcolor{gray!10} 0.0 & \cellcolor{gray!10} N/A & \cellcolor{gray!10} -4.274 & \cellcolor{gray!10} 418.957 
\end{tabular}
\caption{\label{tab:controller_evaluation}Comparison of controller selection and gains on \textbf{Screw} task. \textit{P Motion Gains} specifies the value of the proportional gain for the first $5$ elements of the action space for the motion controller; in order to bias towards efficient screwing, element $6$ was assigned to be $10$ for task-space impedance and $100$ for the others. \textit{D Motion Gains} specifies the value of the derivative gain for all $6$ elements of the action space for the motion controller. \textit{P Force Gains} specifies the value of the proportional gain for all $3$ elements of the closed-loop force controller. \textit{Success Rate} specifies the fraction of episodes that were successful. \textit{Time to Success} specifies the mean number of timesteps required to achieve success. \textit{Reward} specifies the mean reward during each episode. \textit{Joint Torque} specifies the mean joint torque norm during each episode. Each cell is computed from the average of $3$ seeds.}
\end{table*}

\begin{table*}[b]
\centering
\begin{tabular}{lrrcrr}
\textbf{Actions} & \textbf{Observations} & \textbf{Success Rate} & \textbf{Env Steps to Success} & \textbf{Reward} & \textbf{Joint Torque (Nm)} \\ \hline
\multirow{2}{*}{2-DOF (Z, yaw)} & \cellcolor{gray!10} Pose & \cellcolor{gray!10} 0.7734 & \cellcolor{gray!10} \textbf{2308} & \cellcolor{gray!10} -0.1024 & \cellcolor{gray!10} \textbf{1.7378} \\
& Pose, velocity & \textbf{0.7969} & 3658 & \textbf{-0.1004} & 1.7397 \\
\multirow{2}{*}{6-DOF} & \cellcolor{gray!10} Pose & \cellcolor{gray!10} 0.0 & \cellcolor{gray!10} N/A & \cellcolor{gray!10} -0.1602 & \cellcolor{gray!10} 1.7838 \\
& Pose, velocity & 0.0 & N/A & -0.1647 & 1.8008 \\
\end{tabular}
\caption{\label{tab:action_evaluation}Comparison of action spaces on performance of \textbf{Screw} task. For the 2-DOF case, no task-space force/torque was actively generated along the non-active dimensions of the action space. Given the competitive nature of the results in the observation-space assessment, observation spaces were also tested here. \textit{Success Rate} specifies the fraction of episodes that were successful. \textit{Time to Success} specifies the mean number of timesteps required to achieve success. \textit{Reward} specifies the mean reward during each episode. \textit{Joint Torque} specifies the mean joint torque norm during each episode. Each cell is computed from the average of $3$ seeds.}
\end{table*}

\begin{table*}[b]
\centering
\begin{tabular}{lrrr}
\textbf{Baseline} & \textbf{Success Rate} & \textbf{Env Steps to Success} & \textbf{Joint Torque (Nm)} \\ \hline
\rowcolor{gray!10} Linear: $-||d||$ & 0.8359 & 3148 & 1.7295 \\
Exponential: $e^{-||d||}$ & 0.0052 & 3033 & 0.7425 \\
\rowcolor{gray!10} Inverse: $\frac{1}{1 + ||d||}$ & 0.0 & N/A & 0.7270 
\end{tabular}
\caption{\label{tab:reward_evaluation}Comparison of baseline rewards on performance of \textbf{Screw} task. Quantity $||d||$ denotes keypoint distance, defined in \textbf{Section~\ref{sec:screw_policy}}. \textit{Success Rate} specifies the fraction of episodes that were successful. \textit{Time to Success} specifies the mean number of timesteps required to achieve success. \textit{Joint Torque} specifies the mean joint torque norm during each episodes. Due to the distinct range of each baseline reward, mean reward is not provided. Each cell is computed from the average of $3$ seeds.}
\end{table*}

\begin{figure*}
    \centering
    \includegraphics[width=\textwidth]{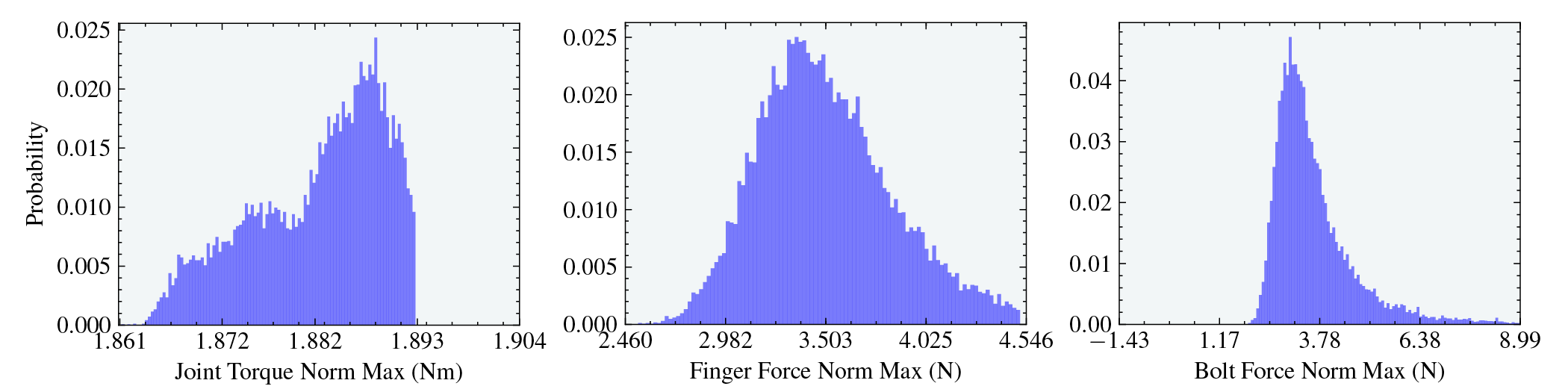}
    \caption{Histograms of joint torques, gripper forces, and bolt forces. Data was collected during inference on the final trained \textbf{Screw} subpolicy over $128$ environments and $1024$ episodes. At each timestep, norms were computed for each environment, and the maximum value was recorded. Histograms show distributions of these values over all timesteps. Extreme outliers were rejected for visualization.}
    \label{fig:validation_force_ranges}
\end{figure*}

\end{document}